%% file: main.tex
\documentclass[10pt]{article} 
\usepackage[accepted]{tmlr}

\input{math_commands.tex}

\usepackage{wrapfig}
\usepackage{hyperref}
\usepackage{url}
\usepackage{amsmath,amsthm,amssymb,bm}
\usepackage{mathtools}
\usepackage{multirow}
\usepackage{booktabs}    
\usepackage{subcaption} 
\usepackage{xcolor} 
\title{HoSNNs: Adversarially-Robust Homeostatic Spiking Neural Networks with Adaptive Firing Thresholds}

\author{\name Hejia Geng \email hejia@ucsb.edu \\
      \addr Department of Electrical and Computer Engineering\\University of California, Santa Barbara
      \AND
      \name Peng Li \thanks{Corresponding author}\email lip@ucsb.edu \\
      \addr Department of Electrical and Computer Engineering\\University of California, Santa Barbara}

\def\month{03}  
\def\year{2025} 
\def\openreview{\url{https://openreview.net/forum?id=UV58hNygne}} 

\begin{document}

\maketitle

\begin{abstract}
While spiking neural networks (SNNs) offer a promising  neurally-inspired model of computation, they are vulnerable to  adversarial attacks. We present the first study that draws inspiration from neural homeostasis to design a  threshold-adapting leaky integrate-and-fire (TA-LIF) neuron model and utilize TA-LIF neurons  to construct the adversarially robust homeostatic SNNs (HoSNNs) for improved robustness. The TA-LIF model incorporates a self-stabilizing dynamic thresholding mechanism, offering a local feedback control solution to the minimization of each neuron's membrane potential error caused by adversarial disturbance.   Theoretical analysis demonstrates favorable dynamic properties of TA-LIF neurons in terms of  the bounded-input bounded-output stability and suppressed time growth of membrane potential error, underscoring their superior  robustness compared with the standard LIF neurons. When trained with weak FGSM attacks (\(\epsilon = 2/255\)), our HoSNNs significantly outperform conventionally trained LIF-based SNNs across multiple datasets. Furthermore, under significantly stronger PGD7 attacks (\(\epsilon = 8/255\)), HoSNN achieves notable improvements in accuracy, increasing from 30.90\% to 74.91\% on FashionMNIST, 0.44\% to 36.82\% on SVHN, 0.54\% to 43.33\% on CIFAR10, and 0.04\% to 16.66\% on CIFAR100.
\end{abstract}

\section{Introduction}
While neural network models have gained widespread adoption across many domains, a glaring limitation of these models has also surfaced — vulnerability to adversarial attacks \citep{szegedy2013intriguing, madry2017towards}. Subtle alterations in the input can trick a well-tuned neural network into producing misleading predictions, particularly for mission-critical applications \citep{chakraborty2018adversarial}. This vulnerability is shared by both artificial neural networks (ANNs) and spiking neural networks (SNNs) \citep{sharmin2019comprehensive, sharmin2020inherent, ding2022snn}, and stands in stark contrast to the inherent robustness of biological nervous systems, prompting interesting questions: Why is the human brain immune to such adversarial noise? Can we leverage biological principles to bolster the resilience of artificial networks? 
\begin{figure*}[ht]
    \centering
    \includegraphics[width=1\textwidth]{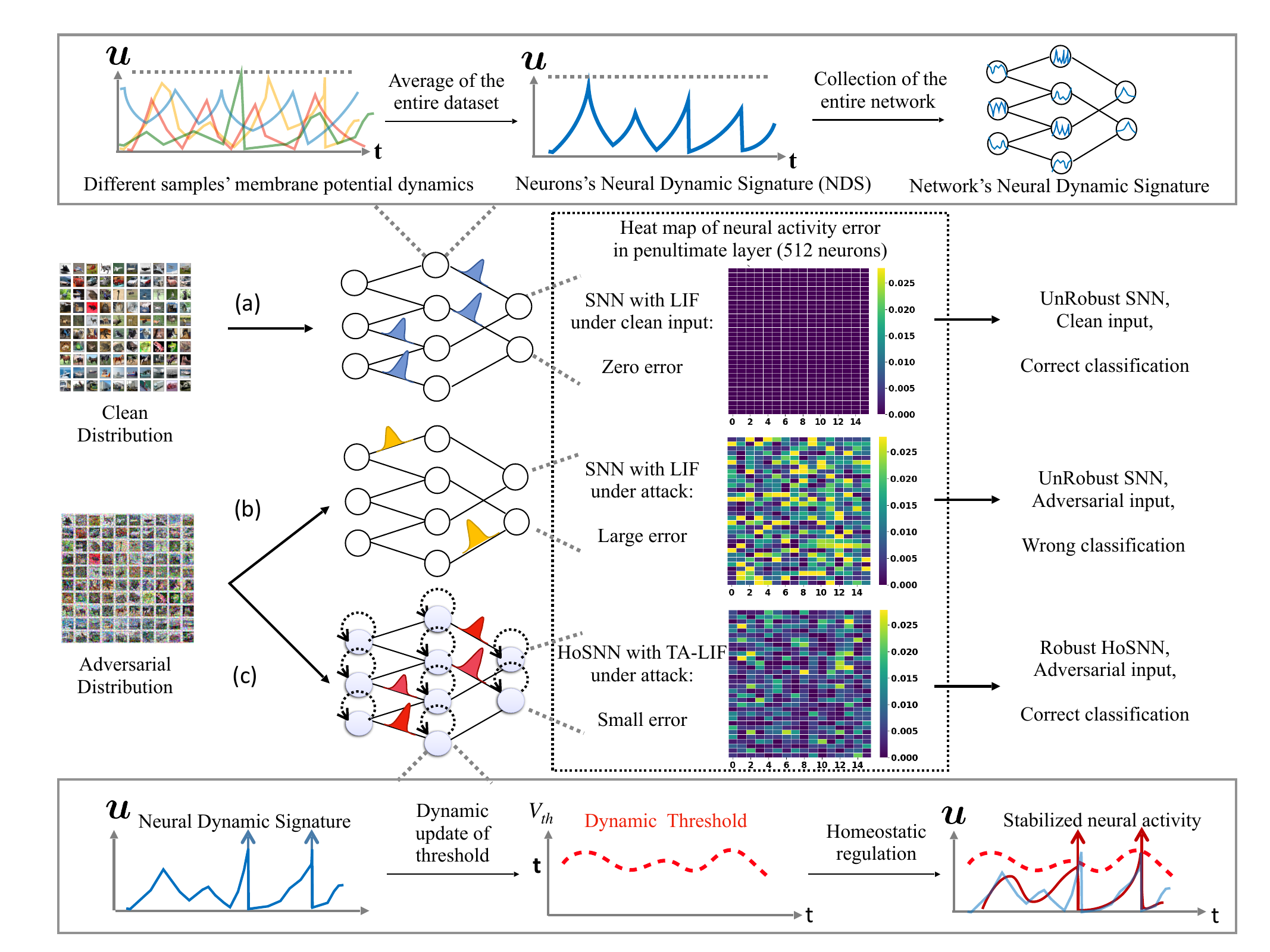}
    \caption{ Proposed threshold-adapting leaky integrate-and-fire (TA-LIF) neuron model and homeostatic SNNs (HoSNNs). (a) We leverage a LIF SNN trained using clean data to collect neural dynamic signatures (NDS) as an anchor for the HoSNN (shown in the box above). (b) Adversarial inputs can cause large membrane potential deviations from the NDS in deep layers of LIF SNNs, leading to incorrect model predictions. (c) Homeostatic dynamic threshold voltage control in HoSNNs anchors neural activity based on the NDS, resulting improved robustness.}
    \label{fig:coreidea}
\end{figure*}

Motivated by these questions, we offer a new perspective that connects adversarial robustness with   homeostatic mechanisms prevalent in living organisms. Homeostasis  maintains essential regulatory variables within a life-sustaining range \citep{bernard1865introduction, cooper2008claude, pennazio2009homeostasis, janig2022integrative}, and is crucial for stabilizing neural activity \citep{turrigiano2004homeostatic}, supporting neurodevelopment \citep{marder2006variability}, and minimizing noisy information transfer \citep{woods2013information, modell2015physiologist}. Although some studies have investigated homeostasis in SNNs, such as the generalized leaky-integrate-and-fire (GLIF) models \citep{Allen, bellec2018long, bellec2020solution, teeter2018generalized}, no prior work  has connected homeostasis with adversarial robustness.

We aim to close this gap by exploring an online biologically-plausible  defense solution based on homeostasis. The proposed approach differs from common practices such as adversarial training in a major way, it explicitly builds a localized neural-level self-adapting feedback mechanism  into the dynamic operation of the proposed HoSNNs. We view the  time-evolving state, i.e., membrane potential \( u_i(t|x )\) of each  spiking neuron $i$ in a well trained network as its representation of the semantics of the received clean input $x$.   Perturbed membrane potential \( u_i(t|x' )\)  resulting from an adversarial input $x'$ corresponds to distorted semantics and can ripple through successive layers to mislead the network output \citep{li2021understanding, rabanser2019failing, nadhamuni2021adversarial, shu2020prepare, fawzi2016robustness, ford2019adversarial, kang2019testing, ilyas2019adversarial}. For a given pair $(x, x')$, we ensure adversarial robustness by minimizing the total induced membrane potential perturbation $E_i(x, x')$. 

We address two practical challenges encountered in formulating and solving this error minimization problem.  Firstly, during inference the clean membrane potential $u_i(t|x)$  reference is unknown. As shown in Fig~\ref{fig:coreidea}(a), we define
Neural Dynamic Signature (NDS), the neuron's membrane potential averaged over a given clean training dataset \( \mathcal{D} \), to provide a reference for anchoring membrane potential. Secondly,  since externally generated attacks are not known \emph{a priori},  it is desirable to suppress the perturbation of each neuron's membrane potential in an online manner as shown in Fig~\ref{fig:coreidea}(b). For this, we propose a new  threshold-adapting leaky integrate-and-fire (TA-LIF) model with a properly designed firing threshold voltage dynamics that serves as a homeostatic control to suppress undesirable membrane potential perturbations, as shown in Fig~\ref{fig:coreidea}(c). We theoretically analyze the dynamic properties of TA-LIF neurons in terms of  the bounded-input bounded-output stability and suppressed time growth of membrane potential error, underscoring their superior  robustness compared with the standard LIF neurons.

\begin{figure*}[ht]
    \centering
    \includegraphics[width=\linewidth]{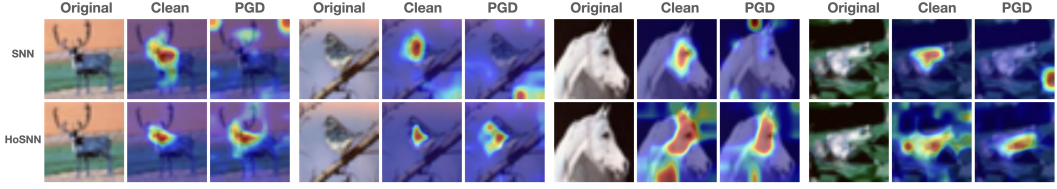}
    \label{fig:talifpic}
    \caption{The heatmaps generated by Grad-CAM \citep{Selvaraju_2019} highlight the regions of the input image that most significantly influence the classification decisions of a standard SNN and proposed HoSNN based on the VGG architecture for a set of CIFAR-10 images. The adversarial images are generated using PGD7 with strength of $\epsilon = 6/255$. HoSNN can still maintain attention to the target object under attack.}
    \label{fig:cifar10}
\end{figure*}

We visualize the working of the proposed HoSNNs versus standard LIF-based SNNs in image classification using several CIFAR-10 images in Figure~\ref{fig:cifar10}. While the adversarial images generated by the Projected Gradient Descent (PGD) \citep{madry2017towards} attack can completely mislead the attention of the LIF SNN, the proposed homeostasis helps the HoSNN focus on parts of the input image strongly correlated with the ground truth class label, leading to significantly improved adversarial robustness as described later.

\section{Background}
\subsection{Adversarial Attacks}
Two notable adversarial attack techniques are the Fast Gradient Sign Method (FGSM) \citep{goodfellow2014explaining}  and  Projected Gradient Descent (PGD) \citep{madry2017towards} method. Let \(x\) be an original input, \(y\) be the true label, \(L(\theta, x, y)\) the loss function with network parameters \(\theta\), and \(\epsilon\) a small perturbation magnitude, FGSM generates a perturbed input, or an adversarial input example  \(x'\) by: $    x' = x + \epsilon \cdot sign[\nabla_x L(\theta, x, y)]  $. PGD is essentially an iterative FGSM. With \(x_n\) as the perturbed input in the $n$-th iteration and \(\alpha\) as the step size, \( Proj_{x+\epsilon}\{ \cdot \} \) as the projection within the \(\epsilon\)-ball of $x$, PGD updates the input by: $x_{n+1} = Proj_{x+\epsilon}\{x_n + \alpha \cdot sign[\nabla_x L(\theta, x_n, y)]\}$. Other gradient-based attacks such as the RFGSM \citep{tramer2017ensemble}, Basic Iterative Method (BIM) \citep{kurakin2018adversarial}, and DeepFool \citep{moosavi2016deepfool} work in a similar fashion, APGD \citep{croce2020reliableevaluationadversarialrobustness}. Beyond gradient-based methods, a significant area of concern is black-box attacks \citep{biggio2018wild}. 

Some recent studies focused on delivering more powerful attacks in SNNs. \cite{bu2023rate} proposed RGA, which leveraged rate-coding in LIF neurons to generate more effective adversarial attacks with a time-extended enhancement. \cite{hao2024threaten} proposed HART, which leveraged a hybrid gradient calculation that simultaneously incorporates rate-based gradients and timing-based temporal gradients on SNNs. 

\subsection{Defense Methods}
Adversarial training is one of the most widely adopted defense methods \citep{madry2017towards}, which retrains a model by using a mixture of clean and adversarial examples. Randomization  \citep{xie2017mitigating} introduces stochasticity during inference  and can  circumvent precise adversarial attacks. The projection technique of  \citep{mustafa2019adversarial} reverts adversarial attacks back to a safer set. Lastly, one may first detect the presence of adversarial attack and subsequently cope with it  \citep{metzen2017detecting}. However, these defense methods are not free of limitations \citep{akhtar2018threat}. Adversarial training relies on precise gradient information and is not biologically plausible \citep{lillicrap2020backpropagation}. Randomization, projection, and detection strategies do not fundamentally address the inherent vulnerabilities of ANNs.

\subsection{Spiking Neuron Models}
SNNs allow for spike-based communication and computation \citep{furber2014spinnaker, gerstner2002spiking, deng2020rethinking} and often leverage  the Leaky Integrate-and-Fire (LIF) model for each neuron \(i\) with the membrane time constant \( \tau_m \):
\begin{equation}
    \label{eq:lif}
    \tau_m \frac{du_i(t)}{dt} = -u_i(t) + I_i(t) - \tau_m s_i(t)V_{th}^i(t),
\end{equation}
 where \(u_i(t)\) is the membrane potential,  \( I_i(t) \triangleq R~\sum_j w_{ij}a_j (t) \) represents the input and is defined as the sum of the  pre-synaptic currents; \( w_{ij} \) represents the synaptic weight from neuron \(j\) to \(i\); \( V_{th}^i(t) \) is the firing threshold of neuron \(i\) at time \(t\). Neuron \(i\)'s postsynaptic spike train is: 
 \begin{equation}
    \label{eq:spike}
    s_i(t) = 
    \begin{cases} 
      +\infty & \text{if } u_i(t) \geq V^i_{th}(t) \\
      0 & \text{otherwise}
    \end{cases}
    = \sum_f \delta(t-t_i^f)
\end{equation}
where $\delta(\cdot)$ is the Dirac function, and \(t_i^f\) is a postsynaptic spike time. With \( \tau_s \) denoting the synaptic time constant, the evolution of the generated postsynaptic current (PSC) \(a_j(t)\) is described by:
\begin{equation}
    \label{eq8}
    \tau_s \frac{da_j(t)}{dt} = -a_j(t) + s_{j}(t)
\end{equation} 
The LIF model uses a constant firing threshold.  Generalized LIF (GLIF) models employ a tunable  threshold with short-term memory, which increases with every emitted output spike, and subsequently decays exponentially back to the baseline threshold \citep{Allen, bellec2018long, bellec2020solution, teeter2018generalized}. However, these models neither consider adversarial robustness nor provide mechanisms to discern "abnormal" from  "normal" neural activity.

\subsection{Spiking Neural Networks Robustness}

While there's been growing interest in spiking neural networks \citep{imam2020rapid,pei2019towards}, empirical studies have demonstrated that SNNs  exhibit similar susceptibilities to adversarial attacks \citep{sharmin2019comprehensive, ding2022snn}. A line of research has explored porting defensive strategies developed for ANNs to SNNs. To improve the robustness of SNNs, \cite{kundu2021hire} proposed a SNN training algorithm jointly optimizing firing thresholds and weights, and \cite{liang2022toward} proposed certified training.  \cite{ding2022snn} enhanced adversarial training by using a Lipschitz constant regularizer. \cite{ozdenizci2023adversarially} introduces an adversarially robust ANN-to-SNN conversion algorithm that initializes the SNN with adversarially pre-trained ANN weights, followed by robust fine-tuning. \cite{liu2024enhancing} improved SNN's adversarial robustness by adding a gradient sparsity regularization term in the loss function. However, these methods have not fully addressed the challenges of ensuring adversarial robustness. Additionally, they are computationally expensive and lack biological plausibility. 

Another research direction has focused on studying the inherent robustness of SNNs not seen in their ANN counterparts, and  factors impacting robustness. \cite{sharmin2020inherent} recognized the inherent resistance of SNNs to gradient-based adversarial attacks. \cite{el2021securing} investigated the impact of key network parameters such as firing voltage thresholds on robustness. \cite{chowdhury2021towards} demonstrated the LIF model's noise-filtering capability. \cite{li2022comparative} explored network inter-layer sparsity. \cite{xu2022securing} examined the effects of surrogate gradient techniques on white-box attacks. While these studies have shed light on  aspects of SNNs  relevant to robustness,  effective defense strategies are yet to be developed.   

Similar to ours, some studies improve robustness from stability and biological and rational perspectives. \cite{ding2024robust} enhanced adversarial robustness by reducing the mean square of perturbations in the last neuron layer. Inspired by Stochastic Gating Mechanisms, \cite{Ding_Yu_Huang_Liu_2024} introduced randomness into spike transmission by simulating the probabilistic opening and closing of synaptic and ion channel gates. The fundamental differences between our work and theirs are: 1.We adjust the threshold in a passive and online manner instead of deliberately adding additional optimization loss terms to the loss function as \citep{ding2024robust}, which significantly reduces the additional calculations for training and inference. 2.Our method does not introduce additional randomness as \citep{Ding_Yu_Huang_Liu_2024}. Thus, our method can retain high clean accuracy and ensures that the robustness does not come from gradient obfuscation \citep{athalye2018obfuscated}.

\section{Method}
 We present the first study that draws inspiration from neural homeostasis to design a threshold-adapting leaky integrate-and-fire (TA-LIF) neuron model and utilize TA-LIF neurons  to construct the adversarially robust homeostatic SNNs (HoSNNs) for improved robustness. 
\subsection{Adversarial Robustness  as a Membrane Potential Error Minimization Problem}
In a well-trained network, we view the time-evolving state, i.e., membrane potential \( u_i(t|x )\) of each  spiking neuron $i$, over $T$ timesteps as its representation of the semantics of the received clean input $x$. {An adversarial input $x' = x+ \delta x,\; s.t.\; \delta x< \epsilon$, where $\epsilon$ is the attack budget (strength), and $\delta x$ is a carefully crafted adversarial noise}, may lead to a perturbed membrane potential \( u_i(t|x' )\), which  corresponds to distorted semantics and can ripple through successive layers to  mislead the network's decision \citep{li2021understanding, rabanser2019failing, nadhamuni2021adversarial, shu2020prepare, fawzi2016robustness, ford2019adversarial, kang2019testing, ilyas2019adversarial}. 

For a given pair of $(x, x')$, one may ensure adversarial robustness of the network by minimizing the total induced membrane potential perturbation $E_i(x, x')$ of   all $N$ neurons over  $T$ timesteps:
\begin{equation}
    \label{eq:opt1}
    \min E_i(x, x') = \sum_{i = 0}^N\sum_{t = 0}^T \parallel u_i(t|x') - u_i(t|x)\parallel^2
\end{equation}
However, there exist two challenges in formulating and solving \eqref{eq:opt1}. Firstly, since the model is oblivious about the attack, it is impossible to determine how the  adversarial input $x'$ is generated, whether it has a corresponding clean input $x$, and what $x$ is if it exists.  As such, $u_i(t|x)$ is unknown, which serves as a clean reference in \eqref{eq:opt1}. We address this problem by inducing the notion of Neural Dynamic Signature (NDS), the neuron's membrane potential averaged over a given clean training dataset \( \mathcal{D} \), to provide an anchor for stabilizing membrane potential. Secondly,  since externally generated attacks are not known \emph{a priori},  it is desirable to suppress the perturbation of each neuron's membrane potential in an online manner. For this, we propose a new type of spiking neurons, called threshold-adapting leaky integrate-and-fire (TA-LIF) neurons with a properly designed firing threshold voltage dynamics that serves as a homeostatic control to suppress undesirable membrane potential perturbations.  We discuss these two techniques next. 

\subsection{Neural Dynamic Signature (NDS) as an Anchor}
{Eq~\ref{eq:opt1} describes an "ideal" optimization problem. When $E_i(x, x')=0$, all neuron activities under adversarial sample input are the same as normal sample input, this can certainly achieve "adversarial robustness" in theory, but it is impossible to achieve in practice. There are two main limitations:(1) For a trained network, the intensity and type of attack are determined by external attackers, which means that x' has a huge range of variation. (2) For a trained network, the attack sample x cannot be determined in advance, which means that it is impossible to obtain an accurate $u_i(t|x) $. Under the constraints of these two problems, we still hope to adopt the idea of Eq~\ref{eq:opt1}. A feasible approximation is that we use the available average value of the training data set $\mathbb{E}_{x \sim \mathcal{D}}\left[u_i(t|x)\right]$ as an approximation of the unavailable $u_i(t|x) $. }

We utilize a clean training dataset \( \mathcal{D} \) to anchor each spiking neuron $i$. While the membrane potential \( u_i(t|x) \)  shows variability across individual samples \( x \), its expected value over distribution \( \mathcal{D} \) denoted by $u^*_i(t|\mathcal{D})$ can act as a reliable reference as illustrated in Fig~\ref{fig:coreidea}(a):
\begin{equation}
\label{eq:nds_def}
u^*_i(t|\mathcal{D}) \triangleq \mathbb{E}_{x \sim \mathcal{D}}[u_i(t| x)]
\end{equation}
The Neural Dynamic Signature (NDS) of neuron \( i \) is defined as a temporal series vector over $T$ timesteps:  $\bm{u^*_i}(\mathcal{D}) = [u^*_i(t_1|\mathcal{D}), u^*_i(t_2|\mathcal{D}), \cdots,  u^*_i(t_T|\mathcal{D})]$.  
\(\bm{u^*_i}(\mathcal{D})\) captures the average semantic activation across \( \mathcal{D} \). Adversarial perturbations induce input distributional shifts, leading to anomalous activation of out-of-distributional semantics.
\(\bm{u^*_i}(\mathcal{D})\) facilitates the identification of neuronal activation aberrations and further offers an anchor signal to bolster network resilience. 

We also define the network-level NDS, $\mathcal{U}_{\text{NET}}(\mathcal{D}) \triangleq \{\bm{u}_i( \mathcal{D})\}_{i=1}^{N}$, as the collection of NDS vectors in a well-trained LIF-based SNN comprising \( N \) neurons. A densely populated SNN with \( N \) neurons typically has \( O(N^2) \) weight parameters. In contrast, $\mathcal{U}_{\text{NET}}(\mathcal{D})$ scales as \( O(NT) \). Recent algorithms have enabled training of high-accuracy SNNs with short latency operating  over a small  number of time steps, e.g., 5 to 10 \citep{zhang2020temporal}. Consequently, the storage overhead of NDS remains manageable.

\paragraph{Dynamics of NDS} 
While serving as an anchor signal, the dynamics of NDS provides a basis for understanding the property of the proposed TA-LIF neurons.    
As NDS  $u^*_i(t|\mathcal{D})$ of neuron $i$ is derived from the well-trained LIF SNN, we take expectation of the LIF dynamic \eqref{eq:lif} with a static firing threshold $V_{th}$ across the entire training distribution $\mathcal{D}$ while simplifying the dynamics by approximating the effects of firing:
\begin{equation}
    \tau_m \frac{d{u^*_i}(t|\mathcal{D})}{dt} = -{u^*_i}(t|\mathcal{D}) +I^*_i(t|\mathcal{D})- \tau_m r^*_i(\mathcal{D}) V_{th}
    \label{eq:nds_dyn}
\end{equation}
Here, $I^*_i(t|\mathcal{D})$ defines the average current input $\mathbb{E}_{x \sim \mathcal{D}}[I_i(t|\theta, x)]$, and   $r^*_i(\mathcal{D})$ denotes the average firing rate $\mathbb{E}_{x \sim \mathcal{D}}[\int_0^{t_T} \frac{s_i(t|x)}{t_T}dt]$. 

{\paragraph{Hypothesis of NDS}
We hypothesize that in SNNs, the original neural activity within the training set holds reference significance for mitigating adversarial perturbations. This idea finds support in certain works within the Artificial Neural Network (ANN) domain, which suggest that detecting and modifying neuron activation values in feature space can alleviate adversarial issues \citep{silva2020opportunitieschallengesdeeplearning, metzen2017detectingadversarialperturbations, zhang2020interpreting}. Some studies deny this view \citep{carlini2017adversarialexampleseasilydetected, tramèr2022detectingadversarialexamplesnearly, ilyas2019adversarialexamplesbugsfeatures, li2021understanding}. In SNNs, whether membrane potential sequences can effectively address adversarial attack problems remains unclear and needs further investigation.}

{Here we provide an intuitive understanding of NDS's effectiveness. Consider an adversarial sample $x' = x + \delta x$, where the perturbation $\delta x$ represents the adversarial attack. As the strength of the perturbation diminishes ($\delta x \rightarrow 0$), the sample gradually converges to the original clean sample ($x' \rightarrow x$). Consequently, the neural activity of the attacked network, denoted as $f(x')$, will approach the clean neural activity, $f(x)$, i.e., $f(x') \rightarrow f(x)$. This alignment between the attacked and clean neural activity suggests that the network's classification result will likewise revert to the correct prediction, $y' \rightarrow y$. Therefore, if we can identify a method to reduce the discrepancy between the neural activity of the attacked network and that of the clean network, we may be able to mitigate the impact of adversarial attacks on the classification outcome.}

\subsection{Threshold-Adapting Leaky Integrate-and-Fire (TA-LIF) Neurons}
\subsubsection{Membrane Potential Error Minimization with NDS}
Instead of examining the deviation of membrane potential $u_i(t| x')$ caused by the adversarial input $x'$ from the unknown $u_i(t|x)$,  we define a new error signal $e_i(t|x') \triangleq u_i(t| x') - u_i^*(t|\mathcal{D})$, and replace the optimization problem of \eqref{eq:opt1} by a more practical membrane potential error minimization problem while optimizing the dynamically changing firing threshold $V^{i}_{th}(t|x')$ of each neuron $i$:
\begin{equation}
\label{eq:opt2}
\min_{\mathbf{V^i_{th}} \in \mathbb{R}^T} E_i(x') = \sum_{t=0}^T e_i(t|x')^2  \triangleq \sum_{t=0}^T (u_i(t| x') - u_i^*(t|\mathcal{D}))^2
\end{equation}
where $\mathbf{V^i_{th}} = [V^{i}_{th}(t_1|x'), V^{i}_{th}(t_2|x'), \cdots, V^{i}_{th}(t_T|x')]$.
The intrinsic parameter of  firing threshold has a critical role in neuronal dynamics and spiking firing. Adapting the firing threshold can mitigate  the effects of adversarial noise, and offer an online homeostatic mechanism for minimizing $E_i(x')$, which is potentially generalizable  across various attack strengths and types.

\subsubsection{Error Minimization of TA-LIF Neurons as a Second-Order Homeostatic Control System}
Subtracting \eqref{eq:nds_dyn} from \eqref{eq:lif} gives the following dynamics of the error $e_i(t)$:
\begin{equation}
    \tau_m \frac{de_i(t|x')}{dt} = -e_i(t|x') +\Delta I_i(t|x') - \tau_m[r_i(x')V^i_{th}(t|x') -r^*_i(\mathcal{D})V_{th}]
    \label{eq:err_dyn}
\end{equation}
 where $\Delta I_i(t|x') \triangleq I_i(t| x')-I_i^*(t|\mathcal{D}) $, and $ r_i(x')  \triangleq \int_0^{t_T} \frac{s_i(t|x')}{t_T} dt $ represents the average firing rate under the adversarial input $x'$. Differentiating \eqref{eq:err_dyn} with respect to time and incorporating the threshold dynamics yields:
\begin{equation}
    \tau_m \frac{d^2e_i(t)}{dt^2} + \frac{de_i(t)}{dt} + r_i \tau_m \frac{dV_{th}^{i}(t)}{d t}  = \varepsilon(t),
    \label{eq:err_dyn2}
\end{equation}
where $\varepsilon(t) \triangleq \frac{d\Delta I_i(t|x')}{dt}$, and we omit the notational dependencies on $x'$  for clarity. Importantly, \eqref{eq:err_dyn2} characterizes the error dynamics $e_i(t)$ as a second-order control system influenced by the external disturbance $\varepsilon(t)$  with $\frac{dV_{th}^{i}(t)}{d t}$ serving as the control term. 

We seek to solve the membrane potential error minimization problem in  \eqref{eq:opt2}  by designing a  control scheme that leads to proper error dynamics based on the second-order error system of \eqref{eq:err_dyn2}.
To this end, we utilize control signal $\frac{dV_{th}^{i}(t)}{d t}$ to provide a negative feedback control to suppress $e_i(t|x')$: 
\begin{equation}
    \label{eq:vth}
    \frac{dV_{th}^{i}(t|x')}{d t} = \theta_i e_i(t|x') = \theta_i[u_i(t| x') - u_i^*(t|\mathcal{D})], 
\end{equation}
where $\theta_i$ is a neuron-level learnable parameter, dictating the pace of firing threshold adjustment.  
 \paragraph{TA-LIF model.}  \eqref{eq:lif}, \eqref{eq:spike}, and \eqref{eq:vth} together delineate the proposed TA-LIF model. 
We construct a homeostatic SNN (HoSNN) using TA-LIF neurons, where each TA-LIF neuron  maintains its unique \(\theta_i\) and \(V_{th}^i(t)\). 
To extract precise semantic information from \(\mathcal{D}\), we collect the NDS for the HoSNN from a well-trained LIF-SNN with identical network configurations. \(\theta_i\) and network weights $W$ of the HoSNN can be jointly optimized using a training algorithm such as backpropagation.  

\begin{figure}[ht]
    \centering
    \begin{subfigure}[b]{0.48\textwidth}
        \centering
        \includegraphics[width=\linewidth]{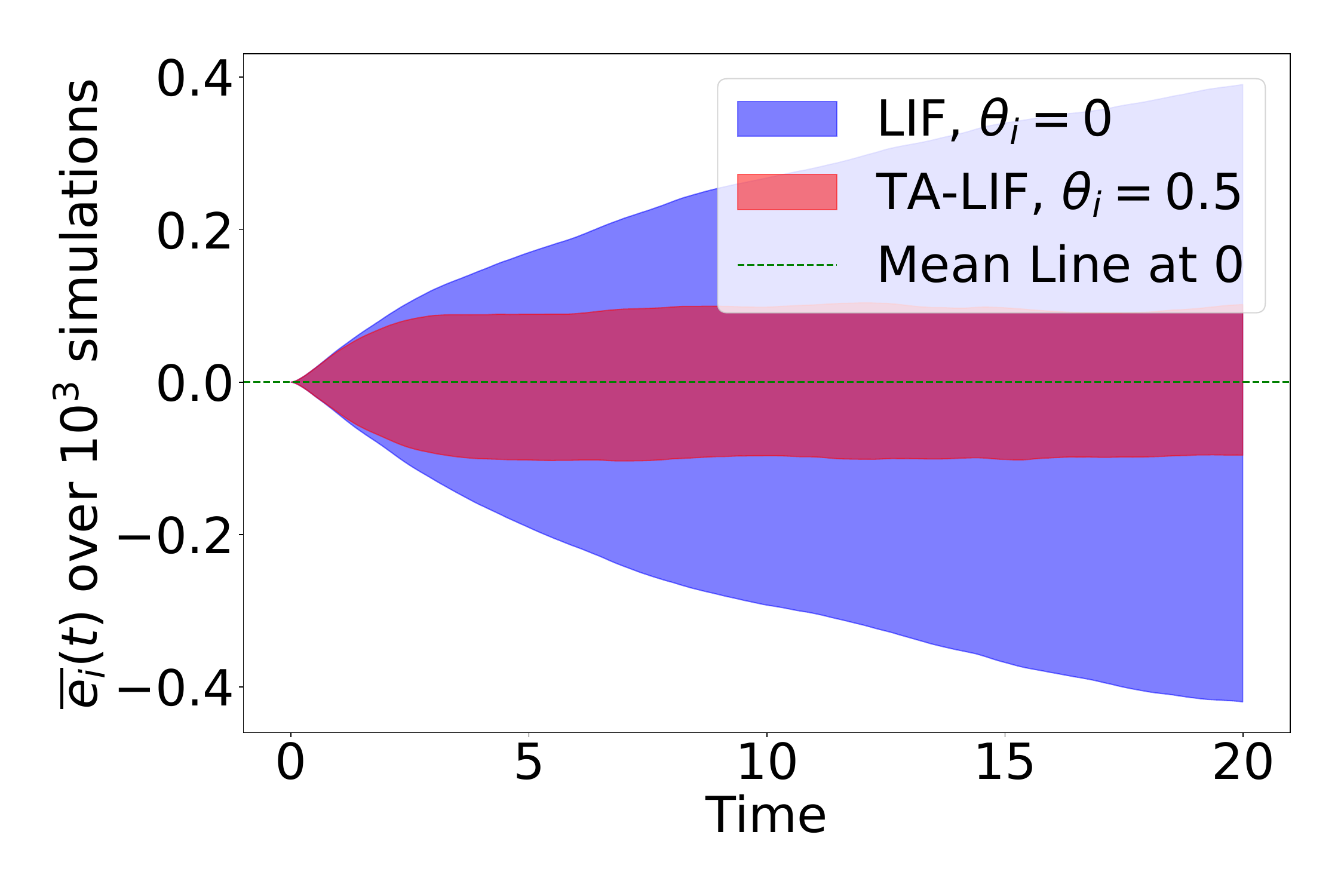}
        \caption{{Numerical simulations of \eqref{eq:err_dyn2}, with parameters $\tau_m=1,\;r=1$}.}
    \end{subfigure}
    \hfill
    \begin{subfigure}[b]{0.48\textwidth}
        \centering
        \includegraphics[width=\linewidth]{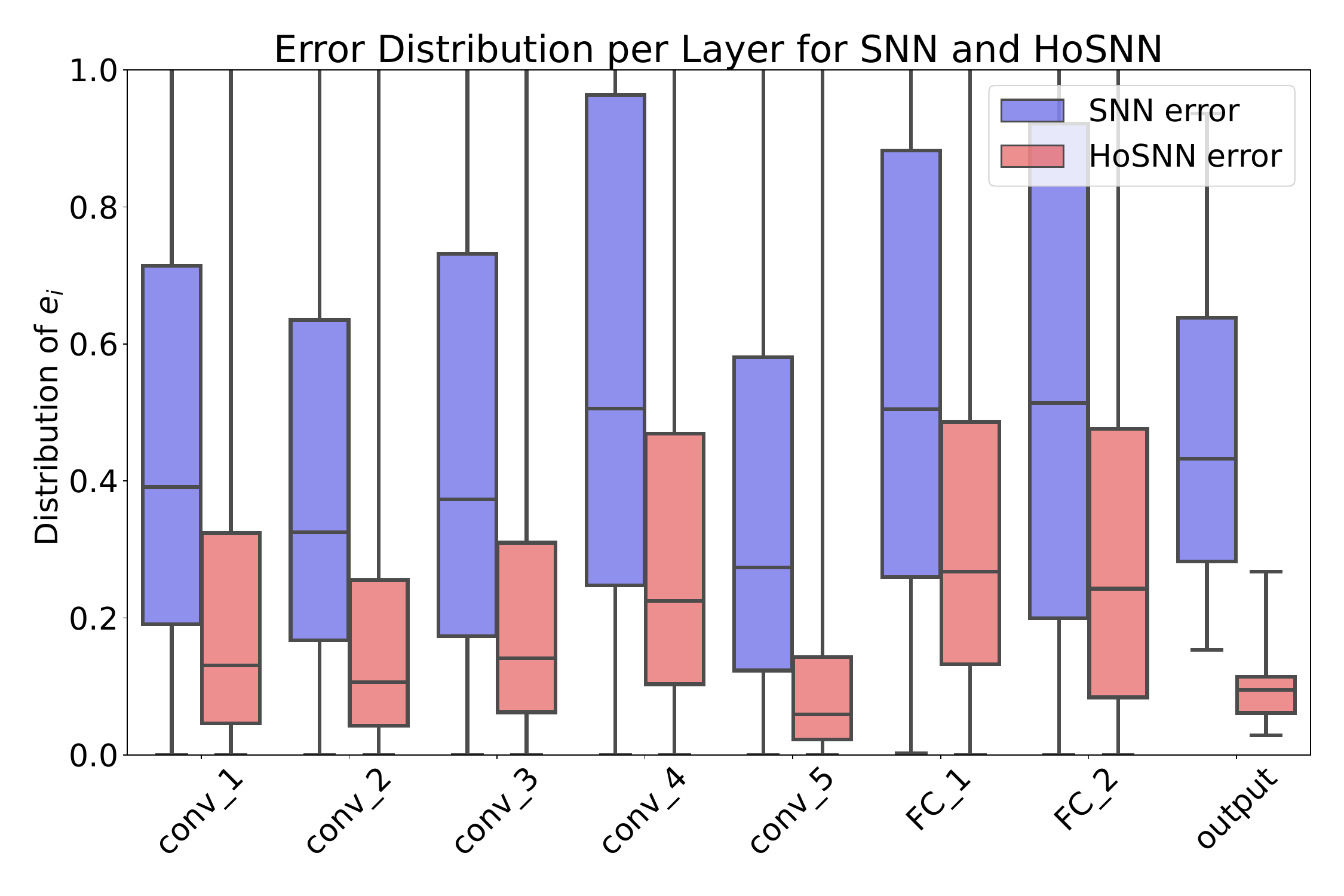}
        \caption{{Box plot of the post-synaptic current relative error distribution per layer of a SNN and a HoSNN.}}
    \end{subfigure}
    \caption{(a) Numerical simulations of \eqref{eq:err_dyn2} show that TA-LIF can well suppress the growth of error with time. (b) Box plot of the post-synaptic current relative error distribution per layer of a SNN and a HoSNN, trained with FGSM adversarial training with $\epsilon = 2/255$  and attacked by a same CIFAR10 black-box PGD7 dataset with $\epsilon = 8/255$.}
    \label{fig:combined_figures}
\end{figure}

During inference with the optimized \(\theta_i\), \(V^i_{th}(t)\) is adapted in an unsupervised manner according to \eqref{eq:vth}. Intuitively, if a TA-LIF neuron $i$ shows a abnormal increased activation relative to the reference NDS, \(V^i_{th}(t)\) would be stepped up to suppress the increase in membrane potential. Conversely, if the neuron is abnormally inhibited, \(V^i_{th}(t)\) would be tuned down. 
Figure~\ref{fig:combined_figures}(a) compares the LIF and TA-LIF models via numerical simulation of \eqref{eq:err_dyn2}, showing the growth of error \( e_i(t) \) over time.  We set the $\varepsilon(t)$ as Gaussian white noise $\xi(t) \sim \mathcal{N}(0, 1)$ and repeated the simulation 1000 times. The membrane potential error of TA-LIF (red area) is significantly smaller than that of LIF (blue area), revealing TA-LIF's dynamic robustness under noisy input perturbations.

The feedback control in \eqref{eq:vth} offers a straightforward means to implement homeostasis, which in turn enhances the adversarial resilience of the proposed HoSNNs. Furthermore, this homeostatic control exhibits two favorable dynamic properties presented next, underscoring its relevance in solving the membrane potential error minimization problem of  \eqref{eq:opt2} as a feedback control solution.   

\subsubsection{Theoretical Dynamic Properties of TA-LIF Neurons}
We highlight key properties of the  TA-LIF dynamics from two perspectives: bounded-input bounded-output (BIBO) stability of membrane potential error and suppressed time growth of error in comparison with the standard LIF model. See \textbf{Appendix A} for complete derivation of these properties.
\paragraph{BIBO Stability} 
If a system is BIBO stable, then the output will be bounded for every input to the system that is bounded.
The characteristic equation and its roots of the proposed second-order TA-LIF dynamics \eqref{eq:err_dyn2} incorporating the homeostatic control \eqref{eq:vth} are:
\begin{equation}
\label{eq:bibo}
    \tau_m s^2 + s + r_i{\tau_m}{\theta_i} = 0, \ s_{1,2} = \frac{-1 \pm \sqrt{1 - 4 r_i {\tau_m^2}{\theta_i}}}{2\tau_m}
\end{equation}
For a second-order system to be BIBO, the roots of its characteristic equation must be a negative real or have a negative real part, which is clearly the case for the TA-LIF model with both $\theta_i$ and $r_i >0$,  affirming the BIBO stability of the TA-LIF model. 
This signifies that when the adversarial input perturbation  $\varepsilon(t)$ is bounded, the deviation of the TA-LIF neuron's membrane potential $e_i(t)$ is also bounded, showing the good control of error under various attack intensities.

\paragraph{Suppressed Time Growth of Membrane Potential Error} To analyze the evolution of membrane potential error induced by injected input perturbations over time, we follow the common practice \citep{gerstner2014neuronal,abbott1993asynchronous, brunel2000dynamics, renart2004mean} to approximate \( \Delta I(t) \) in \eqref{eq:err_dyn2} as a Wiener process, representing small, independent, and random perturbations. Consequently, the driving force $\varepsilon(t)$ on the right of \eqref{eq:err_dyn2}  can be approximated by white noise \( \xi(t) \) with zero mean and variance \( \sigma^2 \). By the theory of stochastic differential equations \citep{kloeden1992stochastic}, this leads to the following mean square error for the LIF and TA-LIF models, respectively:
\begin{equation}
        \text{LIF}: \frac{dV^i_{th}}{dt} = 0 \implies  \langle e_i^2(t)\rangle  = O(\sigma^2t) = \frac{\tau_m^2 \sigma^2}{2} \left(t-\tau_m + \tau_m e^{- t / \tau_m}\right) 
    \label{eq:lifnoise} 
\end{equation}
\begin{equation}
    \text{TA-LIF}: \frac{dV^i_{th}}{dt} = \theta_i e_i  \implies
    \langle e_i^2(t)\rangle = O(\sigma^2 ) = \frac{\tau_m \sigma^2}{2r_i \theta_i}\left[1-e^{\frac{- t}{2\tau_m}}\left(\cos \left(\omega_1 t\right)+\frac{\sin \left(\omega_1 t\right)}{2\omega_1\tau_m}\right)\right] 
    \label{eq:talifnoise}
\end{equation}

where $\omega_1=\sqrt{{r_i}{\theta_i}-\frac{1}{4\tau^2_m}}$.
Importantly, the mean square error of the TA-LIF neuron is $O(\sigma^2 )$ and does not grow with time while that of LIF neurons grows unbounded with time. The suppression of time growth of membrane potential error by the TA-LIF model underscores its superiority over the LIF model in terms of adversarial robustness.

\subsection{Homeostatic SNNs (HoSNNs)}
We further introduce the homeostatic SNNs (HoSNNs), which deploy TA-LIF neurons as the basic compute units to leverage their noise immunity. Architecturally, HoSNNs can be constructed by adopting typical connectivity such as dense or convolutional layers, with two learnable parameters:  synaptic weights \( \boldsymbol{W} \) and threshold dynamics parameter \( \boldsymbol{\theta} \) per \eqref{eq:vth}. We extract the network-level NDS $\mathcal{U}_{\text{NET}}(\mathcal{D})$ from a LIF-based SNN with identical architecture  well-trained on the clean data distribution $\mathcal{D}$. The HoSNNs optimization problem can be described as:
\begin{equation}
          \boldsymbol{W^*}, \boldsymbol{\theta^*} =\arg\min_{\boldsymbol{W}, \boldsymbol{\theta}} \sum_{ \{x,y\}} \mathcal{L}_{\text{train}}( x, y\ |\ \boldsymbol{W},  \boldsymbol{\theta}, \mathbf{V^*_{th}}(x))   \label{eqn:hossn1}
\end{equation}
\begin{equation}
      \text{s.t.}  \mathbf{V^*_{th}}(x)  = \arg\min_{\mathbf{V_{th}}} \mathcal{L}_{\text{mem}}(\boldsymbol{W},  \boldsymbol{\theta}, \mathbf{V_{th}}(x),  \mathcal{U}_{\text{NET}}(\mathcal{{D}})|x)
      \label{eqn:hossn2}
\end{equation}
where \( \mathit{x} \) and \( \mathit{y} \) are an input/label pair; $\mathcal{L}_{\text{train}}(\cdot)$ is the loss over a training dataset that can include clean/adversarial examples, or a combination of the two; $\mathcal{L}_{\text{mem}} \triangleq  \sum_{i=0}^N \sum_{t=0}^T  e_i(t|x)^2 $ is the sum of all neurons'  membrane potential error in \eqref{eq:opt2}. In practice, $\mathcal{L}_{\text{mem}}$ is optimized online by the homeostatic control of firing threshold during the forward process. $\mathcal{L}_{\text{train}}(\cdot)$ is optimized by gradient during the backward process, for which any backpropagation based training algorithm  such as BPTT \citep{neftci2019surrogate, wu2018spatio}, BPTR \citep{lee2020enabling}, or TSSL-BP \citep{zhang2020temporal} can be applied to optimize the network based on \eqref{eqn:hossn1}. 
{
\subsection{Complexity Analysis of HoSNN}
In a standard Spiking Neural Network (SNN), the number of neurons is \( n \), with weights \( w = O(n^2) \) and training time \( T \). For Higher-order Spiking Neural Networks (HoSNNs), additional space is required to store the learnable parameters \( \theta_i = O(n) \), but this storage cost is negligible compared to the network weights. The primary additional cost in training a HoSNN is training the baseline SNN first. The extra time due to the TA-LIF model’s dynamic threshold calculation is minimal (<10\%). Therefore, the overall training time for a HoSNN is about \( 2T \). The dynamic threshold adjustment during inference has negligible impact, so the inference cost of a trained HoSNN is almost identical to that of an SNN.
There are some challenges in the training of HoSNN. (1) \( \theta_i \) Initialization: Initializing \( \theta_i \) too large (> 1) can destabilize training. It is recommended to set \( \theta_i \) between 0 and 0.5, with smaller values needed for complex datasets. (2) Optimizer Choice: HoSNNs converge slower in later training stages, especially in adversarial settings. Using the Adam optimizer with a cosine decay learning rate helps achieve similar convergence speeds as SNNs.
}
\section{Experiments}\label{sec:results}
\subsection{Experimental Setup} The proposed HoSNNs are  compared with LIF-based SNNs with identical architecture across four benchmark datasets: Fashion-MNIST (FMNIST) \citep{xiao2017fashion},  Street View House Numbers (SVHN) \citep{netzer2011reading} CIFAR10 and CIFAR100 \citep{krizhevsky2009learning}.  VGG-5/9/11 \citep{simonyan2014very} convolutional neural network (CNN) architectures of different sizes and depths are utilized. Widely used methods including FGSM \citep{goodfellow2014explaining}, RFGSM \citep{tramer2017ensemble}, PGD (iteration = 7, 20, 40) \citep{madry2017towards}, and BIM \citep{kurakin2018adversarial} are used to generate both white-box and black-box attacks. To test stronger attacks, we introduce APGD \citep{croce2020reliable} in our white-box attack, with cross entropy loss ($\text{APGD}_{\text{CE}}$), difference of logits ratio loss ($\text{APGD}_{\text{DLR}}$) and targeted attack ($\text{T-APGD}$). Unless otherwise specified, PGD refers to PGD7. \textbf{The dynamically changing firing thresholds of the HoSNNs are exposed to the attacker and utilized in the gradient calculation when generating white-box attacks.} We independently trained SNNs with the same architecture and used their white-box attacks as the black-box attacks to the HoSNNs.

For each HoSNN,  an LIF-based SNN with an identical architecture is trained on the corresponding clean dataset to derive the NDS. Model training  employs the BPTT learning algorithm (\(T=5\)), leveraging a sigmoid surrogate gradient \citep{xu2022securing, neftci2019surrogate, wu2018spatio}. The learning rate for each $\theta_i$ in \eqref{eq:vth}, which controls the adaptation of the firing threshold of TA-LIF neurons, is set to 1/10 of that for the network weights, ensuring hyperparameter stability during training. We ensure that  $\theta_i$ is non-negative during optimization. 
We train four types of models: SNNs and HoSNNs on a clean dataset and a weak FGSM-based adversarial training dataset, respectively. For FGSM adversarial training, we set the attack budget to \(\epsilon = 2/255\) on FMNIST, SVHN 
 and CIFAR10 as in \citep{ding2022snn} and \(\epsilon = 4/255\) on CIFAR100 as in \citep{kundu2021hire}. For iterative attacks (PGD \& BIM), we adopt parameters \(\alpha=2.5*\epsilon/steps\) and \(steps=7, 20, 40\) in accordance with \citep{ding2022snn}. Mode experimental settings are in the Appendix C.

\subsection{Adversarial Robustness under White-box Attacks}
\paragraph{Robustness without adversarial training}
Table \ref{combined_table1} compares HoSNNs and SNNs intrinsic resilience without adversarial training. Both types of network are trained exclusively on the clean dataset, and then subjected  to $\epsilon = 8/255$ white-box adversarial attacks. The HoSNNs show consistently higher accuracy than the SNN counterparts under all four datasets and four attacks. For example, on CIFAR-10, the HoSNN significantly improves  accuracy from 20.86\% to 54.76\% under FGSM, from 0.54\% to 15.32\% under PGD7, from 0.69\% to 10.35\% under APGD with cross entropy loss, from 4.44\% to 16.02\% under T-APGD.

\paragraph{Robustness with adversarial training} In Table~\ref{combined_table2}, we evaluate the enhanced robustness of HoSNNs under adversarial training. We train both the SNNs and HoSNNs using FGSM adversarial training and then expose them to $\epsilon = 8/255$ white-box attacks. The results show a significant boost in robustness for HoSNNs when using  adversarial training. Furthermore, the HoSNNs noticeably outperform the SNNs trained using the same FGSM adversarial training. For example, on CIFAR10, the HoSNNs improve the accuracy of the corresponding SNN to 63.98\% from 37.93\% under FGSM attack, and to 43.33\% from 12.42\% under PGD7 attack, from 8.23\% to 38.89\% under APGD-CE attack. On CIFAR100, the HoSNN improves the accuracy to 16.66\% from 8.82\% under PGD7 attack, from 7.75\% to 12.55\% under APGD-CE attack.
\begin{table*}[ht]
    \centering
    \footnotesize
    \begin{tabular}{clllllllllll}
    \toprule
        Dataset & Net  & Clean & FGSM & RFGSM & BIM7  & PGD7 & PGD20 & PGD40 & $\text{APGD}_{\text{CE}}$ & $\text{APGD}_{\text{DLR}}$ & $\text{T-APGD}$   \\
     \hline
    \multirow{2}{*}{\shortstack{Fashion\\MNIST}} 
    & $\times$ & 92.92  & 56.01 & 70.02 & 38.85 & 30.90 & 28.73 & 27.88  & 23.22 & 40.25 & 39.67\\
    & $\checkmark$  & $\mathbf{92.96}$ & $\mathbf{65.36}$ & $\mathbf{76.10}$ & $\mathbf{49.02}$ & $\mathbf{35.79}$ & $\mathbf{32.99}$ & $\mathbf{32.63}$ & $\mathbf{37.78}$ & $\mathbf{61.01}$ & $\mathbf{50.57}$ \\
    \hline
    \multirow{2}{*}{\shortstack{SVHN} }
    & $\times$ &  $\mathbf{95.51}$  & 26.07 & 42.94 & 2.26 & 0.44 & 0.18 & 0.13 & 0.11 & 7.73 & 0.81 \\
    & $\checkmark$   & 93.55  & $\mathbf{44.87}$ & $\mathbf{57.27}$ & $\mathbf{12.91}$ & $\mathbf{4.33}$ & $\mathbf{1.87}$ & $\mathbf{1.47}$ & $\mathbf{3.09}$ & $\mathbf{12.26}$ & $\mathbf{5.32}$\\
    \hline
    \multirow{2}{*}{\shortstack{CIFAR\\10}} 
    & $\times$ &  $\mathbf{92.47}$ & 20.86 & 38.72 & 3.29 & 0.54 & 0.38 & 0.3 & 0.69 & 7.03 & 4.44   \\
    & $\checkmark$    &  92.43 & $\mathbf{54.76}$ & $\mathbf{62.33}$  & $\mathbf{28.06}$ & $\mathbf{15.32}$& $\mathbf{11.58}$ & $\mathbf{10.74}$ & $\mathbf{10.35}$ & $\mathbf{27.39}$ & $\mathbf{16.02}$  \\
    \hline
    \multirow{2}{*}{\shortstack{CIFAR\\100}} 
    & $\times$ & $\mathbf{74.00}$ & 5.74 & 8.94 & 0.10 & 0.04 & 0.01 & 0.00 &0.00 & 0.01 & 0.01   \\
    & $\checkmark$  &  71.98 & $\mathbf{13.48}$ & $\mathbf{12.27}$ & $\mathbf{0.50}$ & $\mathbf{0.19}$ & $\mathbf{0.02}$ & $\mathbf{0.02}$ & $\mathbf{2.55}$ & $\mathbf{0.02}$ & $\mathbf{0.02}$\\
    \bottomrule
    \end{tabular}
    \caption{Training on clean dataset. Whitebox attack results on Fashion-MNIST, SVHN,  CIFAR10 and CIFAR100 under various types of attack with an intensity of $\epsilon = 8/255$. The data on the left and right are based on training using the clean and  weak FGSM datasets, respectively. HoSNNs (denoted as $\checkmark$) provide greater robustness than SNNs (denoted as $\times$) under all attacks and datasets.}
    \label{combined_table1}
\end{table*}

\begin{table*}[ht]
    \centering
    \footnotesize
    \begin{tabular}{clllllllllll}
    \toprule
        Dataset & Net  & Clean & FGSM & RFGSM & BIM7  & PGD7 & PGD20 & PGD40 & $\text{APGD}_{\text{CE}}$ & $\text{APGD}_{\text{DLR}}$ & $\text{T-APGD}$   \\
     \hline
    \multirow{2}{*}{\shortstack{Fashion\\MNIST}} 
    & $\times$  &92.08  & 74.12 & 83.12 & 68.36  & 62.92 & 61.97 & 61.43  & 55.47 & 68.25  & 62.56 \\
    & $\checkmark$  & $\mathbf{92.31}$ & $\mathbf{84.7}$ & $\mathbf{87.99}$ & $\mathbf{79.61}$ & $\mathbf{74.91}$ & $\mathbf{73.53}$ & $\mathbf{73.21}$ & $\mathbf{67.94}$ & $\mathbf{76.24}$ & $\mathbf{74.48}$    \\
    \hline
    \multirow{2}{*}{\shortstack{SVHN} }
    & $\times$ & $\mathbf{93.85}$ & 48.37 & 68.81  & 30.09 & 18.46 & 15.52 & 14.79 & 9.70 & 25.66 & 20.82\\
    & $\checkmark$  &  92.84 & $\mathbf{61.78}$ & $\mathbf{75.60}$ & $\mathbf{48.83}$ & $\mathbf{36.82}$ & $\mathbf{32.24}$ & $\mathbf{30.89}$ & $\mathbf{28.20}$ & $\mathbf{47.89}$ & $\mathbf{43.96}$ \\
    \hline
    \multirow{2}{*}{\shortstack{CIFAR\\10}} 
    & $\times$  & $\mathbf{91.87}$ & 37.93 & 59.50 & 22.31 & 12.42 & 11.03 & 10.63 & 8.23 & 16.8 & 14.62 \\
    & $\checkmark$  & 90.00 & $\mathbf{63.98}$ & $\mathbf{71.07}$  & $\mathbf{52.33}$ & $\mathbf{43.33}$ & $\mathbf{40.97}$ & $\mathbf{40.02}$ & $\mathbf{38.89}$ & $\mathbf{37.94}$ & $\mathbf{41.69}$  \\
    \hline
    \multirow{2}{*}{\shortstack{CIFAR\\100}} 
    & $\times$  &  $\mathbf{68.72}$ & 22.54 & 36.93 & 13.58 & 8.82 & 7.88 & 7.52 & 7.75 & 7.51 & 5.4  \\
    & $\checkmark$  & 64.64 & $\mathbf{26.97}$ & $\mathbf{41.45}$ & $\mathbf{21.09}$ & $\mathbf{16.66}$ & $\mathbf{15.83}$& $\mathbf{15.37}$ & $\mathbf{12.55}$ & $\mathbf{13.66}$ & $\mathbf{10.2}$
    \\
    \bottomrule
    \end{tabular}
    \caption{Training on FGSM dataset ($\epsilon = 2/255$). Whitebox attack results on Fashion-MNIST, SVHN,  CIFAR10 and CIFAR100 under various types of attack with an intensity of $\epsilon = 8/255$. The data on the left and right are based on training using the clean and  weak FGSM datasets, respectively. HoSNNs (denoted as $\checkmark$) provide greater robustness than SNNs (denoted as $\times$) under all attacks and datasets.}
    \label{combined_table2}
\end{table*}

\subsection{Adversarial Robustness under Black-box Attacks}
\begin{figure}[h]
    \includegraphics[width=\linewidth]{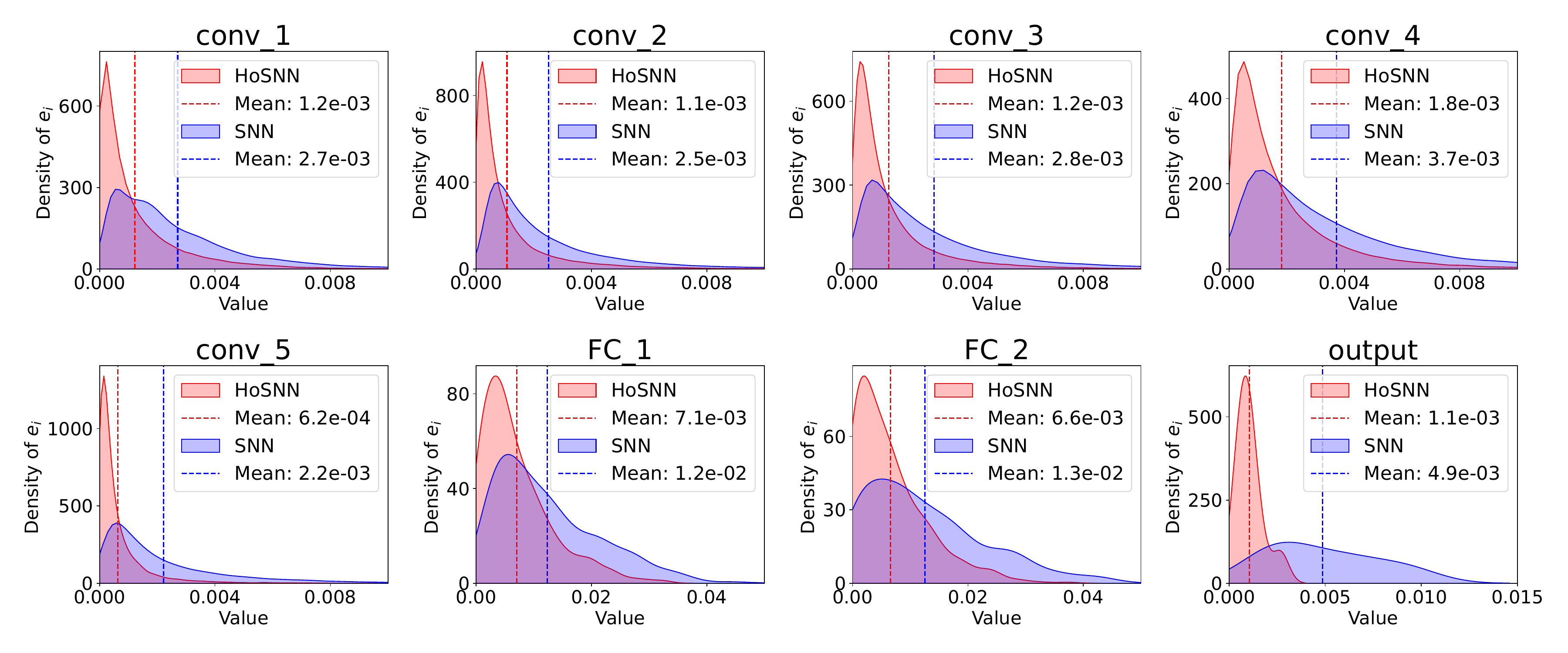}
    \centering
    \caption{Distribution of post-synaptic current relative error of a SNN and HoSNN trained with FGSM adversarial training with $\epsilon = 2/255$  and attacked by black-box PGD7 attack with $\epsilon = 8/255$.}
    \label{fig:distribution}
\end{figure}

\paragraph{Layer-wise Relative Error of Post-synaptic Currents}
Perturbation in membrane potential caused by adversarial inputs can alter the output spike train of each neuron, and the resulting shifts in its post-synaptic current (PSC) can propagate through the successive layers. The PSC $a_i(t)$ is calculated by \eqref{eq8}.  To reveal the source of HoSNNs robustness, we examine the relative error in PSC induced by adversarial attacks.  On CIFAR10, we use the $\epsilon = 8/255$ black-box PGD7 attack to attack SNN and HoSNN trained with the $\epsilon = 2/255$ FGSM adversarial training. For neuron $i$, we record its PSC $a_i(t)$ and $a'_i(t)$ under each pair of clean and adversarial input, respectively, and then calculate the difference of  $e^{PSC}_i(t) \triangleq |a_i(t)-a'_i(t)|$ as an error metric. The layer-wise distributions of the relative error $e^{PSC}_i$ are plotted in Figure~\ref{fig:combined_figures}(b) and Figure~\ref{fig:distribution}. 

The experiment results show that each error distribution of the HoSNN has a significantly reduced mean compared with that of the SNN, and has its probability mass concentrated on low PSC error values, revealing the favorable internal self-stabilization introduced by the proposed homeostasis. Correspondingly,  the HoSNN delivers an accuracy of 76.62\%, significantly surpassing the 46.97\% accuracy of the SNN.

\paragraph{Black Box Robustness}
\begin{wraptable}{r}{0.55\textwidth}
    \centering
    \footnotesize
    \begin{tabular}{cclllllll}
        \toprule
        Dataset & Net & Clean & FGSM & RFGSM & PGD7 & BIM7   \\
        \hline
        \multirow{2}{*}{\shortstack{Fashion\\MNIST}}
        &  $\times$ &92.08 & 66.26 & 80.09 & 74.08 & 73.43 \\
        &  $\checkmark$ & $\mathbf{92.31}$    & $\mathbf{68.31}$ & $\mathbf{80.73}$ & $\mathbf{75.12}$ & $\mathbf{74.09}$  \\
        \hline
        \multirow{2}{*}{\shortstack{SVHN}}
        &  $\times$ &$\mathbf{93.85}$  & 17.37 & 42.64 & 21.16 & 36.70   \\
        &  $\checkmark$  & 92.84  & $\mathbf{19.08}$ & $\mathbf{44.57}$ & $\mathbf{25.97}$ & $\mathbf{40.75}$  \\
        \hline    
        \multirow{2}{*}{\shortstack{CIFAR\\10}}
        & $\times$  & $\mathbf{91.87}$ & 13.48 & 8.79 & 0.11 & 0.31  \\
        & $\checkmark$  & 90.00 & $\mathbf{25.18}$ & $\mathbf{31.11}$ & $\mathbf{13.42}$ & $\mathbf{23.34}$  \\
            \hline
        \multirow{2}{*}{\shortstack{CIFAR\\100}}
        &  $\times$ &$\mathbf{68.72}$ & 12.18 & 17.87 & 6.74 & 17.36  \\
        &  $\checkmark$ & 64.64 & $\mathbf{14.54}$ & $\mathbf{24.32}$ & $\mathbf{16.90}$ & $\mathbf{32.04}$   \\
        \bottomrule
    \end{tabular}
    \caption{SNNs and HoSNNs black-box attack accuracy, trained with FGSM adversarial training and tested by different black-box attack methods with $\epsilon = 32/255$.}
    \label{bbatk}
\end{wraptable}
Table~\ref{bbatk} evaluates the robustness of SNNs (denoted as $\times$) and HoSNNs (denoted as $\checkmark$) against black-box attacks. All models  are trained by weak FGSM adversarial training and tested by $\epsilon = 32/255$ black-box attacks generated using separately trained SNNs with identical architecture. Table~\ref{bbatk} shows that the HoSNNs exhibit significantly stronger black-box robustness than the SNN counterparts. For example on CIFAR-10, the HoSNN  outperforms the traditional SNN under the FGSM and PGD7 attacks with 11.7\% and 13.31\% accuracy improvements, respectively. On CIFAR-100, HoSNN improves the PGD7 accuracy from 6.74\% to 16.90\%. 
\subsection{Checklist for gradient obfuscation} For a detailed check of \textbf{gradient obfuscation} \citep{athalye2018obfuscated, carlini2019evaluating}, we provide comprehensive inspection and data (in Appendix D). HoSNNs pass all five tests as show in Table~\ref{grad_obf}

\begin{table}[h]
    \centering
    \begin{tabular}{lcc}
        \hline 
        Items to identify gradient obfuscation & HoSNN & Experiment \\
        \hline 
        (1) Single-step attack performs better compared  to iterative attacks & \checkmark & Fig~\ref{fig:fgmpgd_compare} and Table~\ref{tab:transposed_fmnist_other_results1}
\\
        (2) Black-box attacks perform better compared to white-box attacks & \checkmark & Fig~\ref{fig:wb_compare} and Table~\ref{tab:transposed_fmnist_other_results2} \\
        (3) Increasing perturbation bound can't increase attack strength & \checkmark & Fig~\ref{fig:ho_compare} and Table~\ref{tab:transposed_fmnist_other_results3}\\
        (4) Unbounded attacks can't reach $100 \%$ success  & \checkmark & Fig~\ref{fig:ho_compare} and Table~\ref{tab:transposed_fmnist_other_results3} \\
        (5) Adversarial example can be found through  random sampling & \checkmark & Fig~\ref{fig:performance}\\
        \hline
    \end{tabular}
    \caption{Checklist for gradient obfuscation}
    \label{grad_obf}
\end{table}

\begin{figure}[h]
            \centering
            \begin{subfigure}{0.245\linewidth}
                \includegraphics[width=\linewidth]{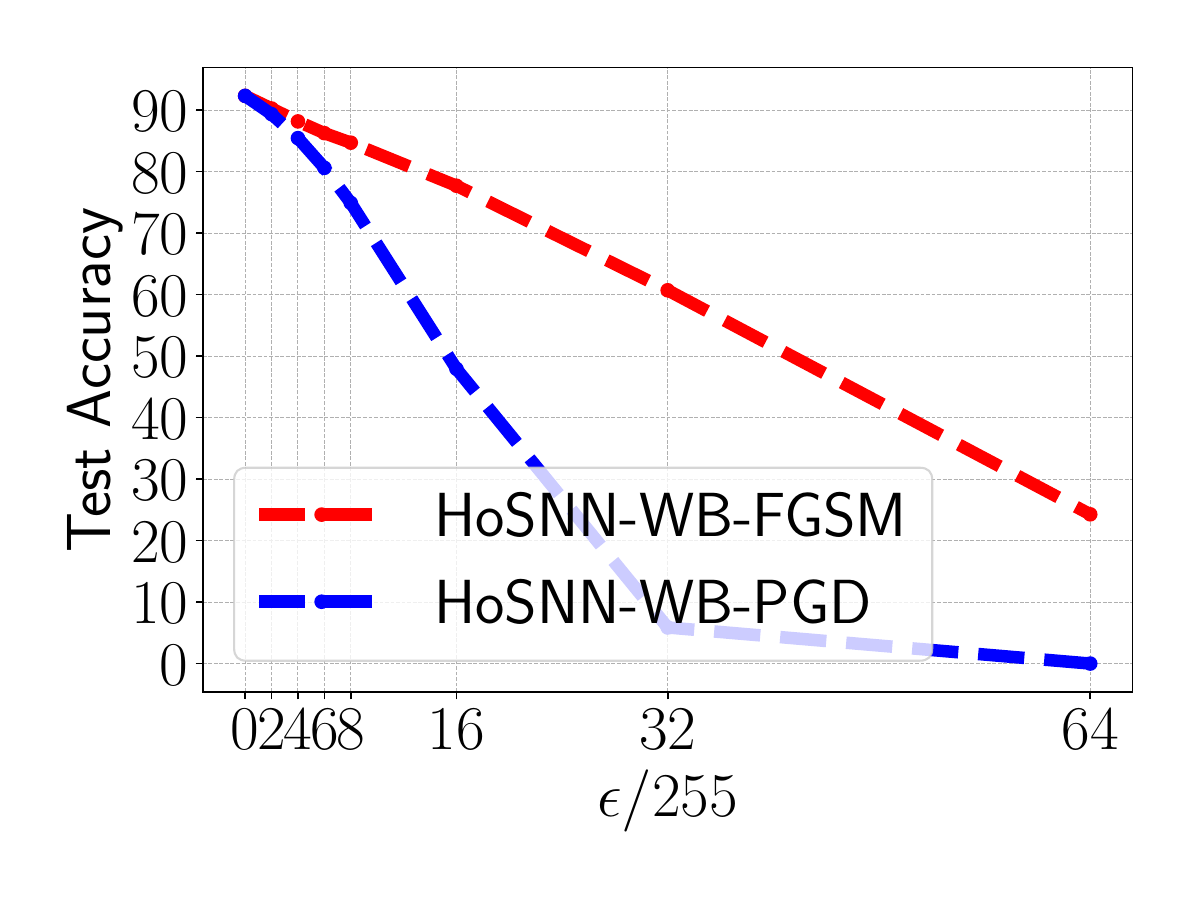}
                \caption{Fashion-MNIST}
            \end{subfigure}
            \begin{subfigure}{0.245\linewidth}
                \includegraphics[width=\linewidth]{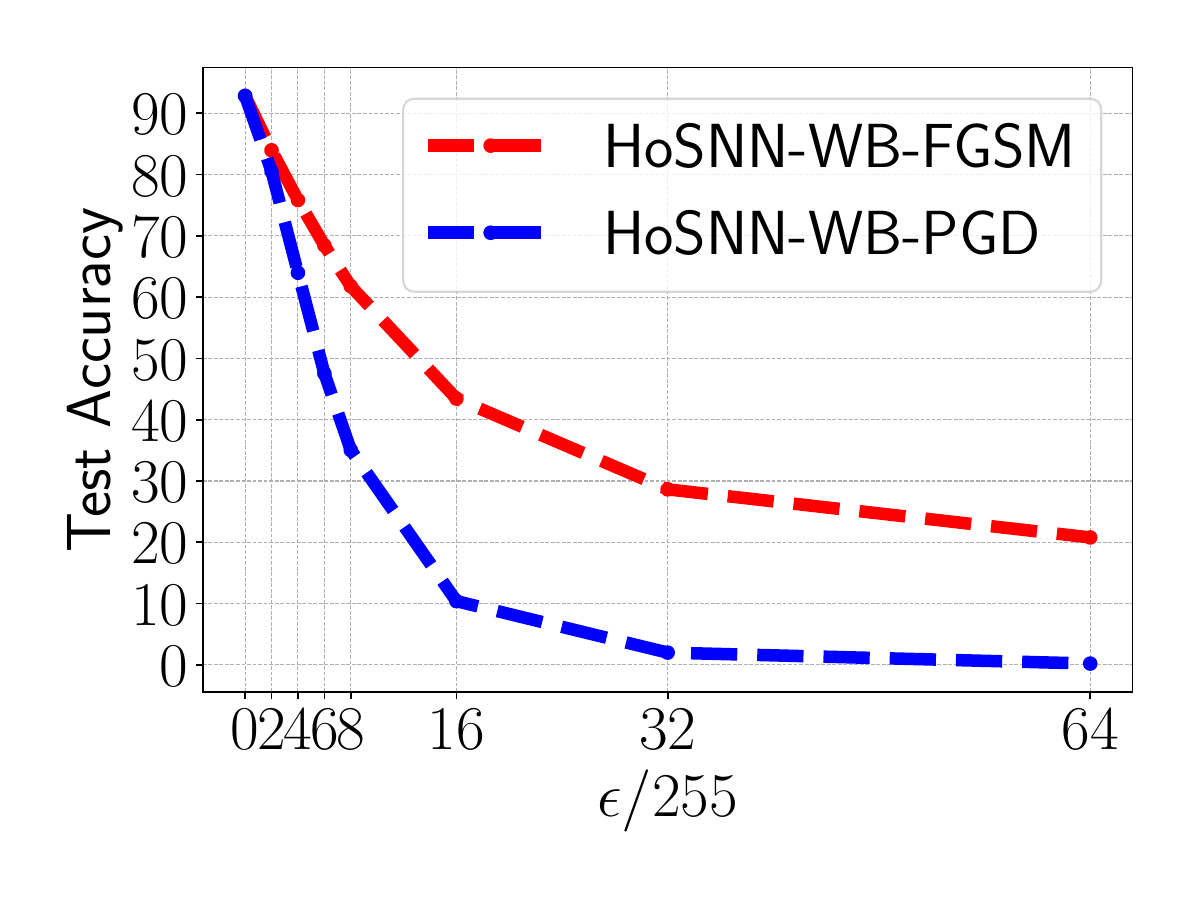}
                \caption{SVHN}
            \end{subfigure}
            \begin{subfigure}{0.245\linewidth}
                \includegraphics[width=\linewidth]{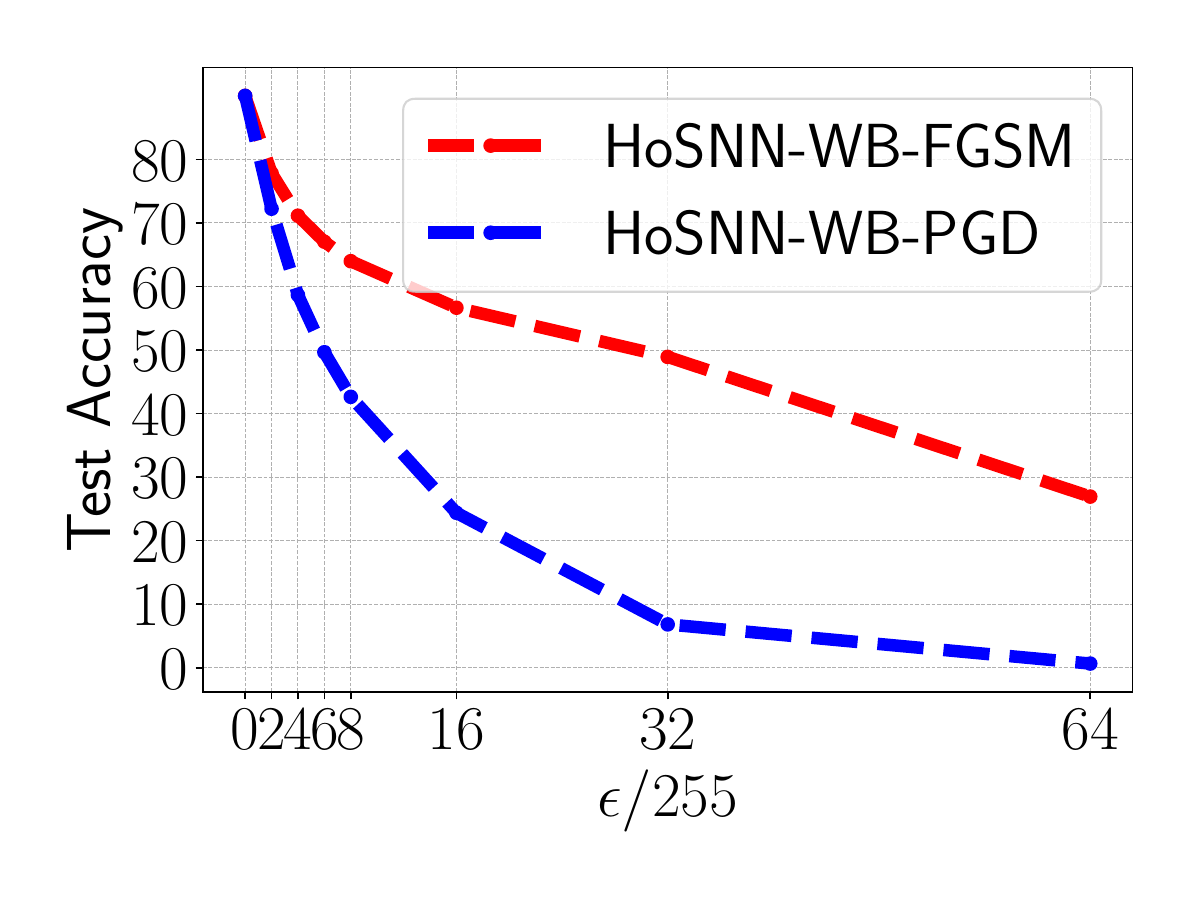}
                \caption{CIFAR-10}
            \end{subfigure}
            \begin{subfigure}{0.245\linewidth}
                \includegraphics[width=\linewidth]{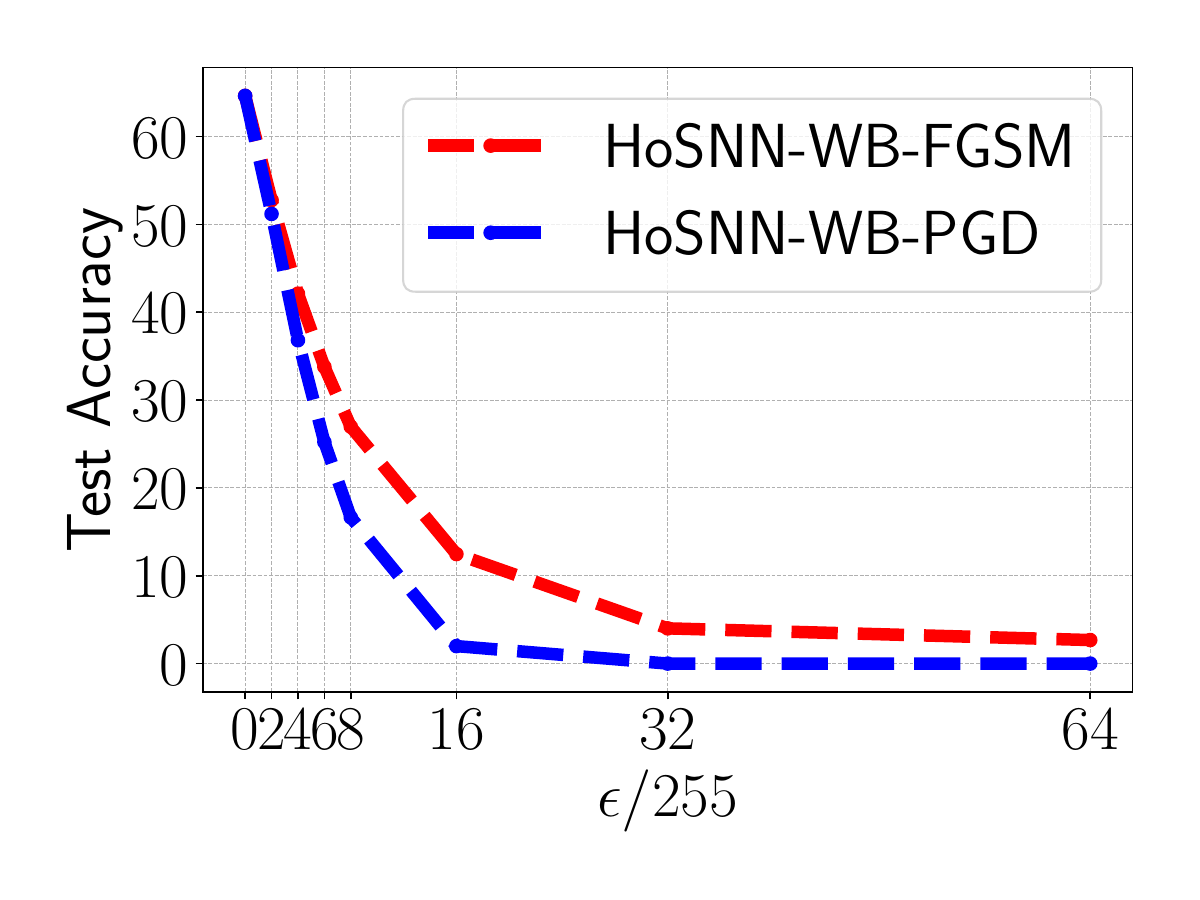}
                \caption{CIFAR-100}
            \end{subfigure}
            \caption{For Test (1). Performance of HoSNN under white-box FGSM and PGD7 attack.}
            \label{fig:fgmpgd_compare}
        \end{figure}

    \paragraph{For Test (1)} We plot the curves of white-box FGSM and PGD7 attacks on four datasets in Figure~\ref{fig:fgmpgd_compare} and Table~\ref{tab:transposed_fmnist_other_results1}, with attack budgets $\epsilon$ from 0 to 64/255 to ensure that the network can be completely fooled. The red curve is the accuracy under FGSM, while the blue curve is PGD7. From the Figure~\ref{fig:fgmpgd_compare} we can confirm that all iterative attacks are much stronger than single-step attack.

\begin{figure}[h]
            \centering
            \begin{subfigure}{0.245\linewidth}
                \includegraphics[width=\linewidth]{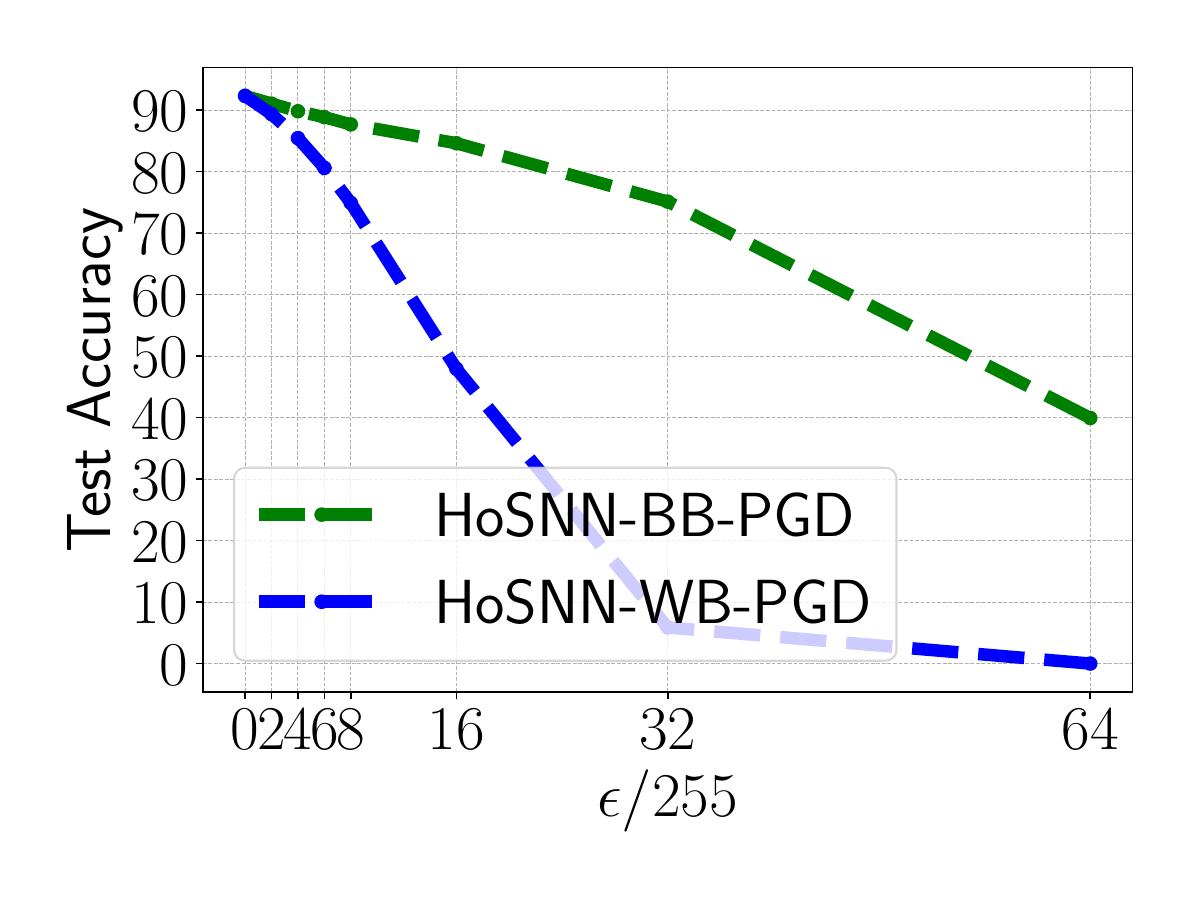}
                \caption{Fashion-MNIST}
            \end{subfigure}
            \begin{subfigure}{0.245\linewidth}
                \includegraphics[width=\linewidth]{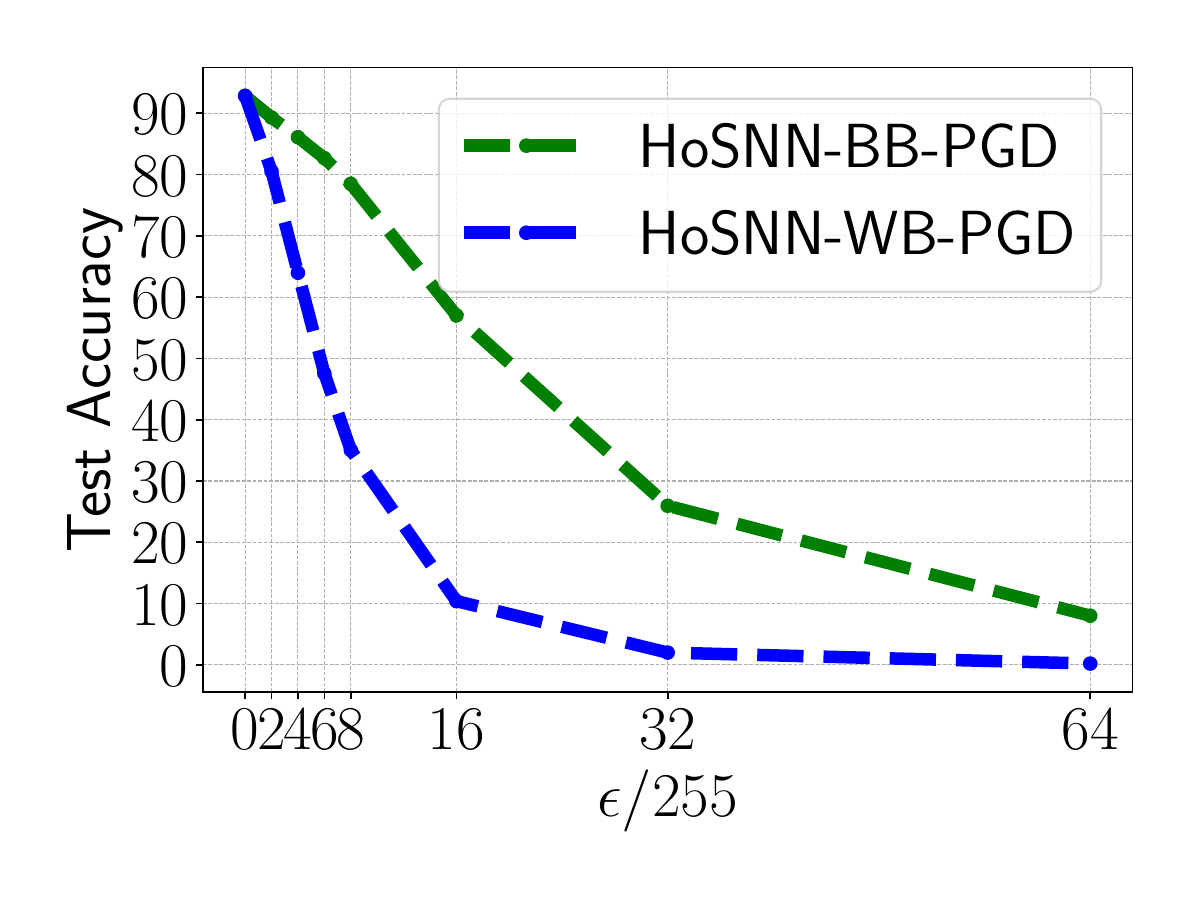}
                \caption{SVHN}
            \end{subfigure}
            \begin{subfigure}{0.245\linewidth}
                \includegraphics[width=\linewidth]{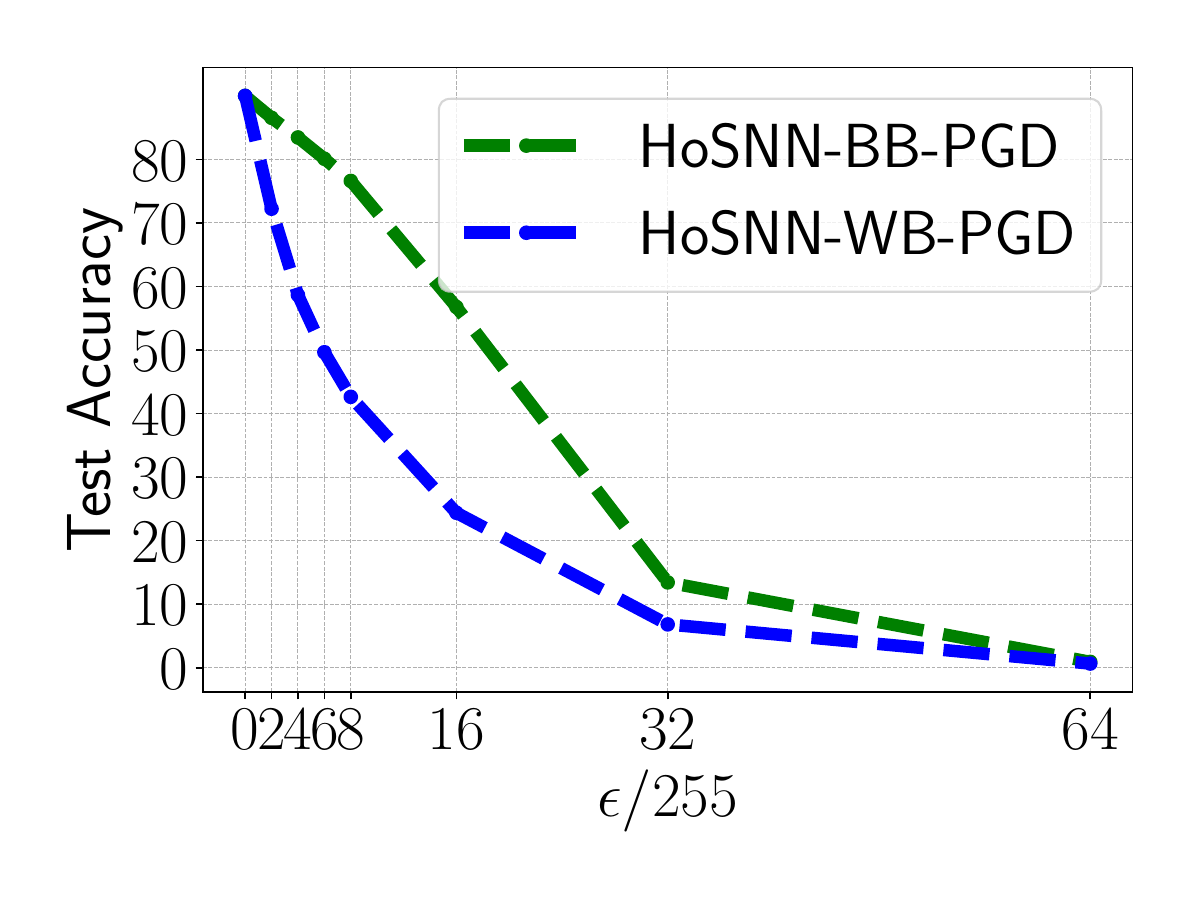}
                \caption{CIFAR-10}
            \end{subfigure}
            \begin{subfigure}{0.245\linewidth}
                \includegraphics[width=\linewidth]{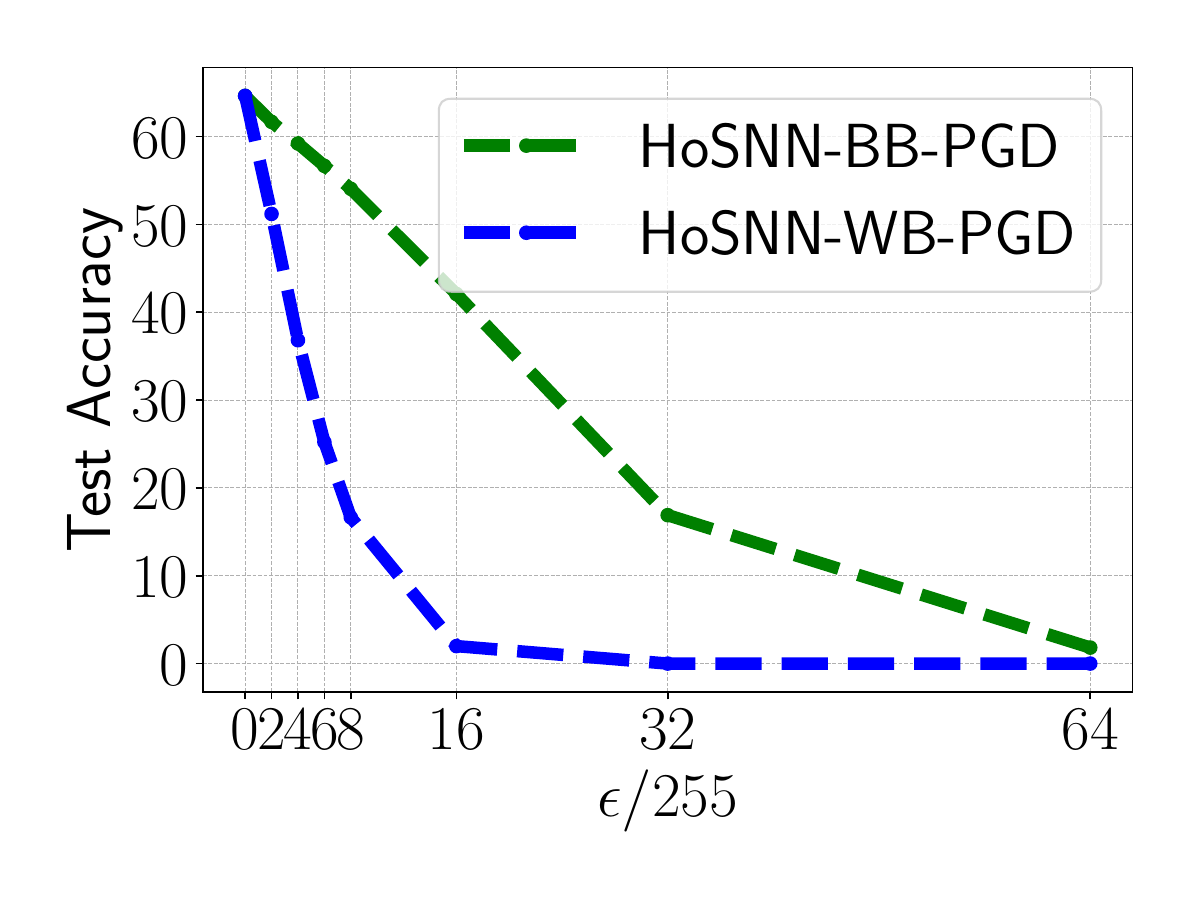}
                \caption{CIFAR-100}
            \end{subfigure}
            \caption{For Test (2). Performance of HoSNN under white-box and black-box and PGD7 attack.}
            \label{fig:wb_compare}
        \end{figure}

    \paragraph{For Test (2)} We plot the curves of white-box and black-box PGD7 attacks on four datasets in Figure~\ref{fig:wb_compare}  and Table~\ref{tab:transposed_fmnist_other_results2}, with attack budgets $\epsilon$ from 0 to 64/255 to ensure that the network can be completely fooled. The green curve is the accuracy under black-box PGD7 attack, while the blue curve is under white-box. From the Figure~\ref{fig:wb_compare} we can confirm that all white-box attacks are much stronger than black-box. 
   
\begin{figure}[h]
            \centering
            \begin{subfigure}{0.245\linewidth}
                \includegraphics[width=\linewidth]{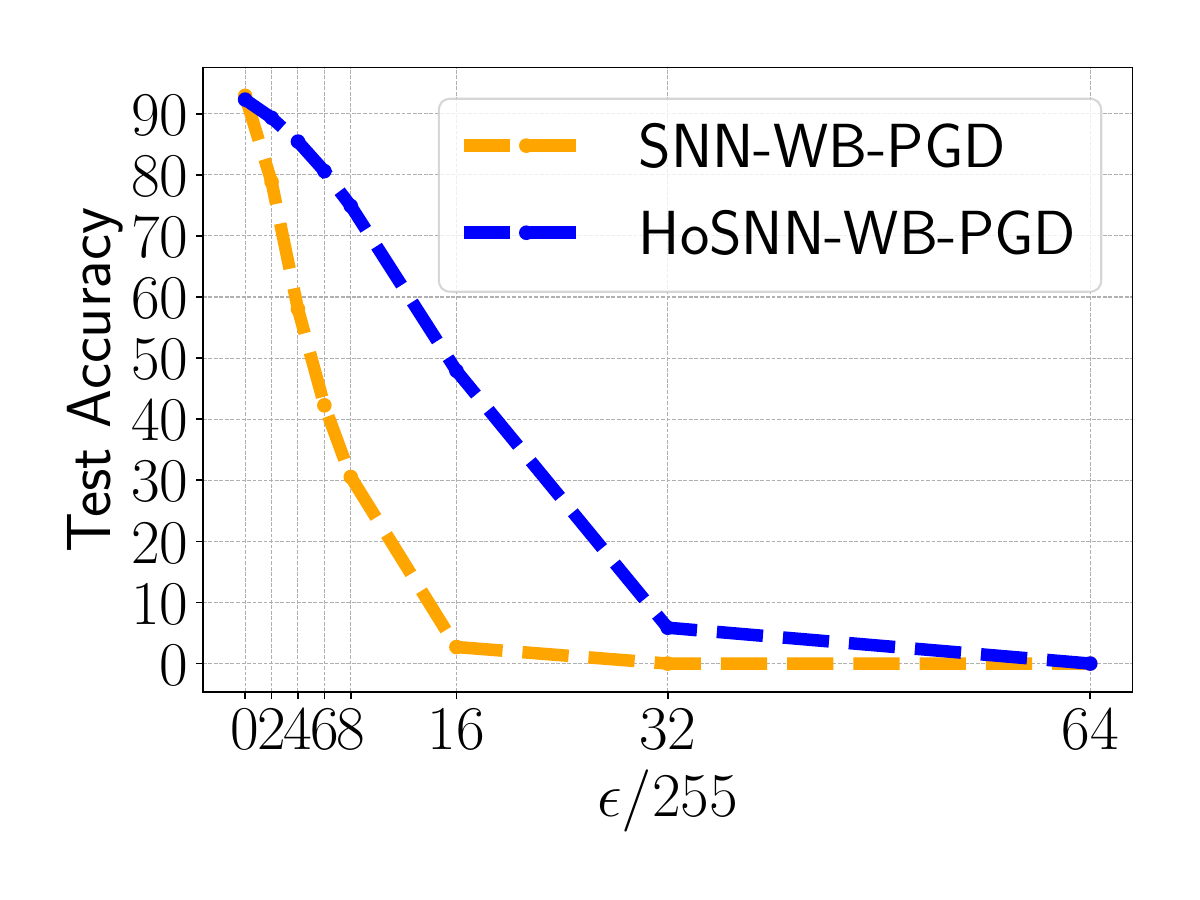}
                \caption{Fashion-MNIST}
            \end{subfigure}
            \begin{subfigure}{0.245\linewidth}
                \includegraphics[width=\linewidth]{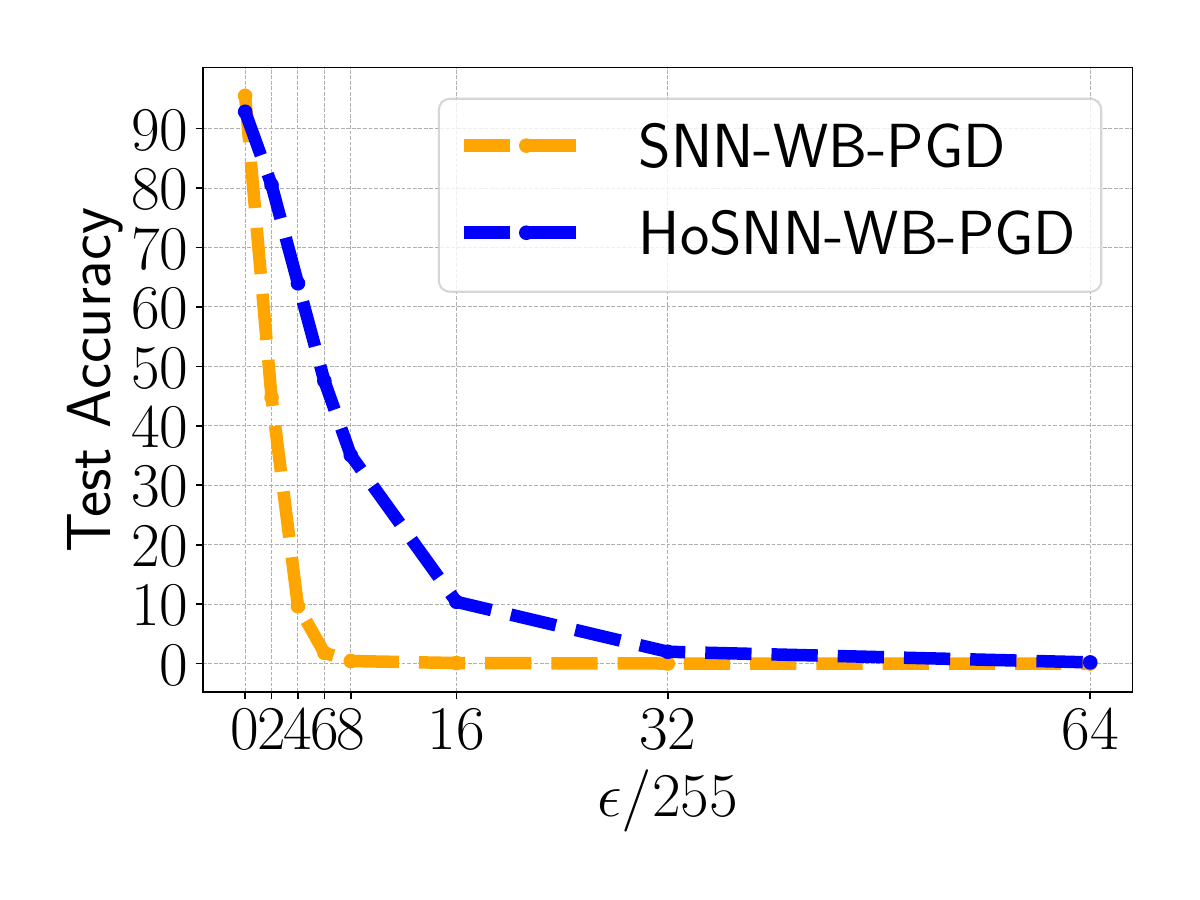}
                \caption{SVHN}
            \end{subfigure}
            \begin{subfigure}{0.245\linewidth}
                \includegraphics[width=\linewidth]{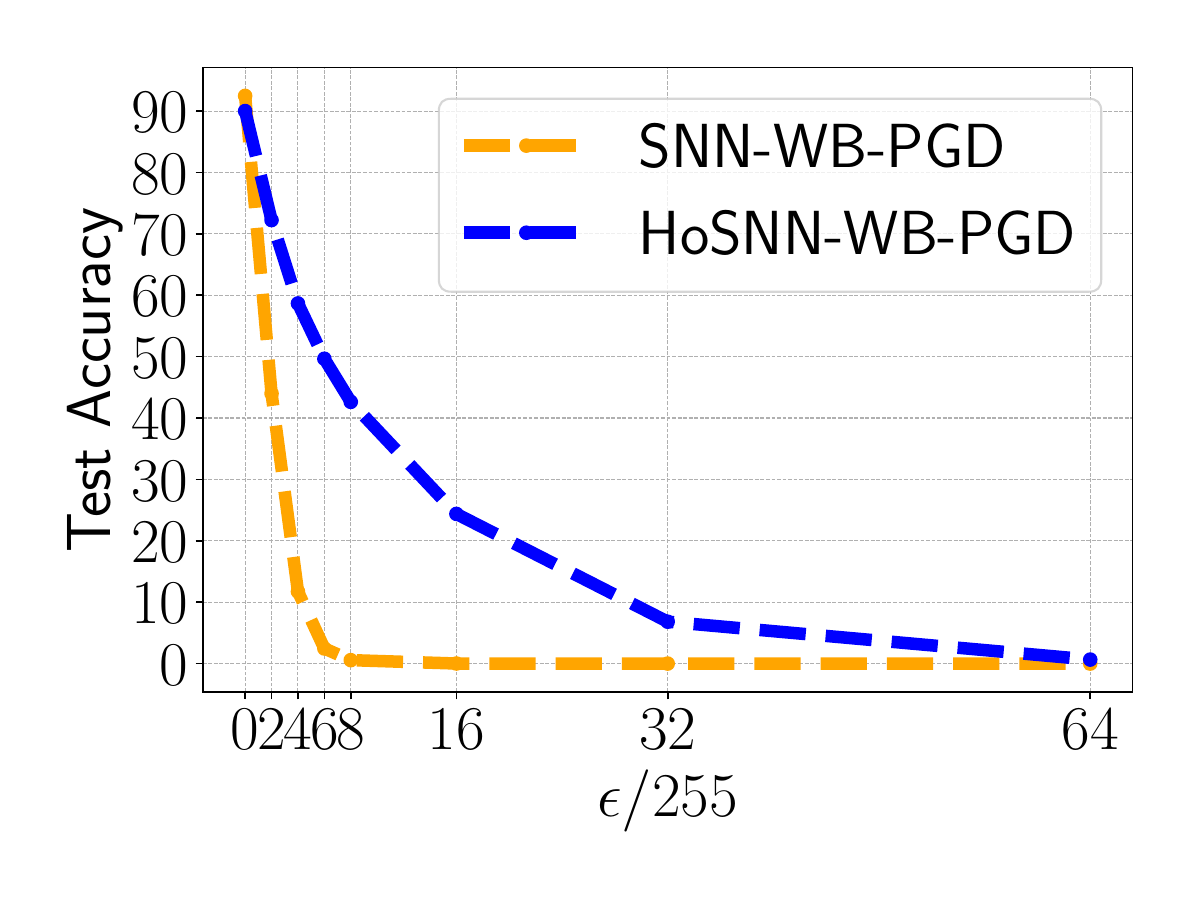}
                \caption{CIFAR-10}
            \end{subfigure}
            \begin{subfigure}{0.245\linewidth}
                \includegraphics[width=\linewidth]{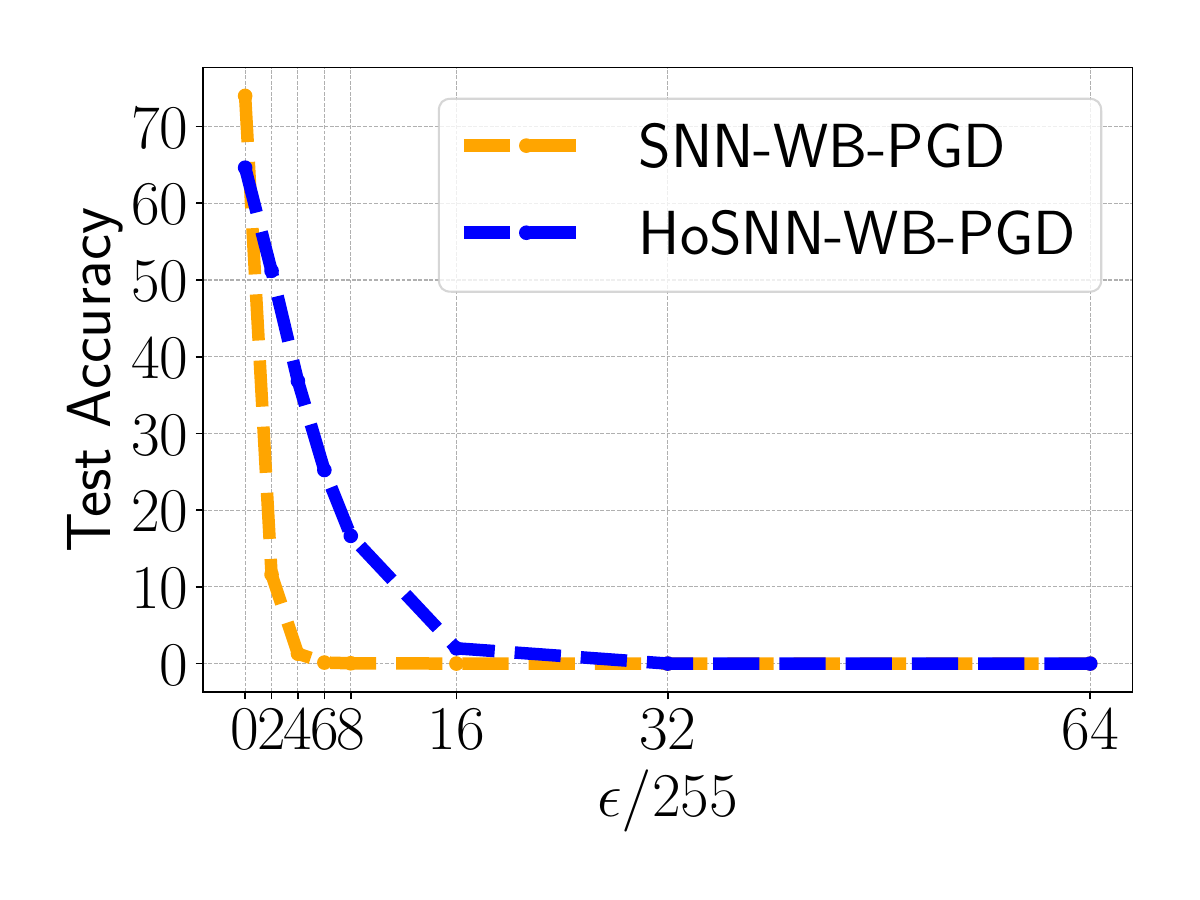}
                \caption{CIFAR-100}
            \end{subfigure}
            \caption{For Test(3) and (4). Performance of HoSNN under larger white-box PGD7 attack.}
            \label{fig:ho_compare}
        \end{figure}

 \paragraph{For Test (3) \& (4)} We increasing perturbation bound can't increase attack strength \& Unbounded attacks can't reach $\sim 100 \%$ success, we plot the curves of white-box and black-box PGD7 attacks on four datasets in Figure~\ref{fig:ho_compare}  and Table~\ref{tab:transposed_fmnist_other_results3}, with attack budgets $\epsilon$ from 0 to 64/255 to ensure that the network can be completely fooled. The blue curve is HoSNN's accuracy under white-box PGD7 attack, while the yellow curve is SNN's baseline. From the Figure~\ref{fig:ho_compare} we can confirm that as perturbation bound increasing HoSNN's accuracy is decreasing and all unbounded attacks reach $\sim 100 \%$ success.

\begin{figure}[h]
            \centering
            \begin{subfigure}{0.245\linewidth}
                \includegraphics[width=\linewidth]{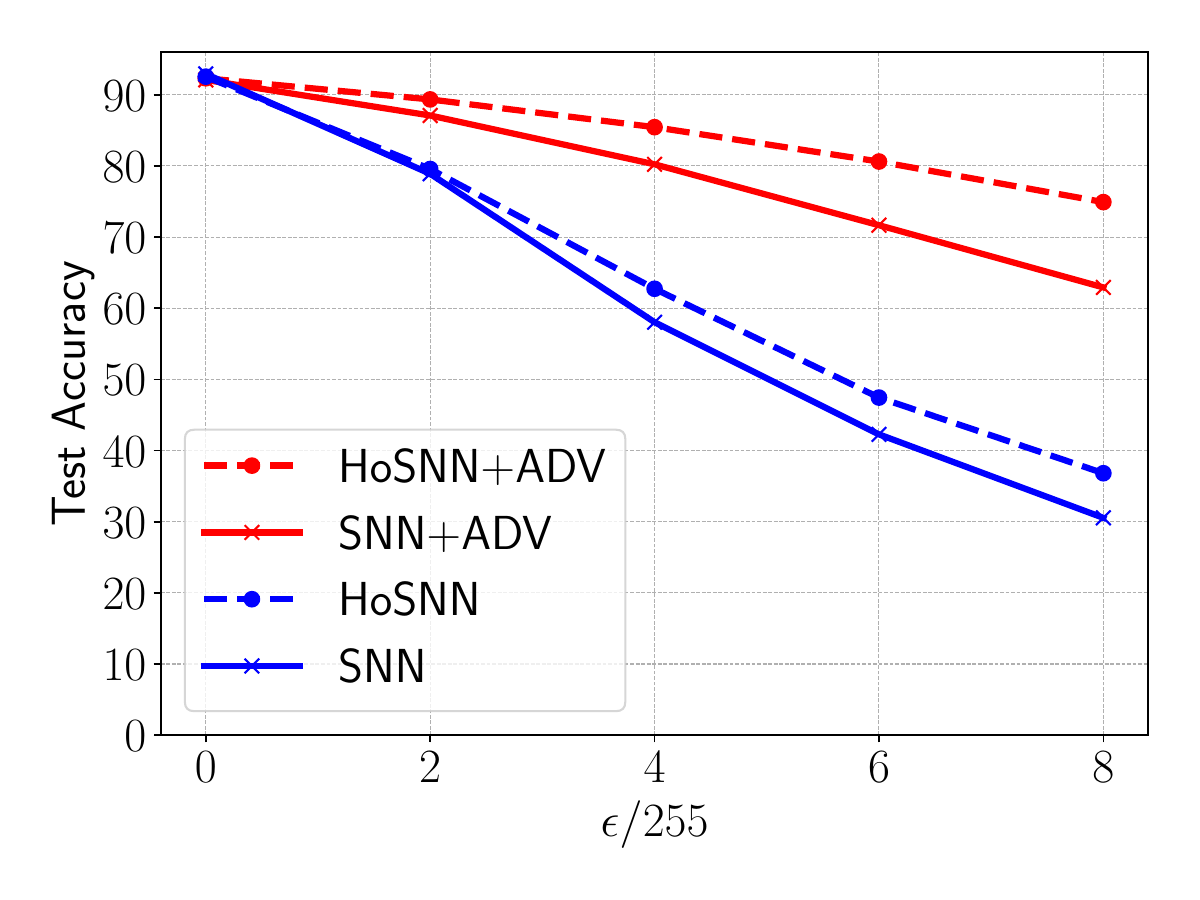}
                \caption{Fashion-MNIST}
            \end{subfigure}
            \begin{subfigure}{0.245\linewidth}
                \includegraphics[width=\linewidth]{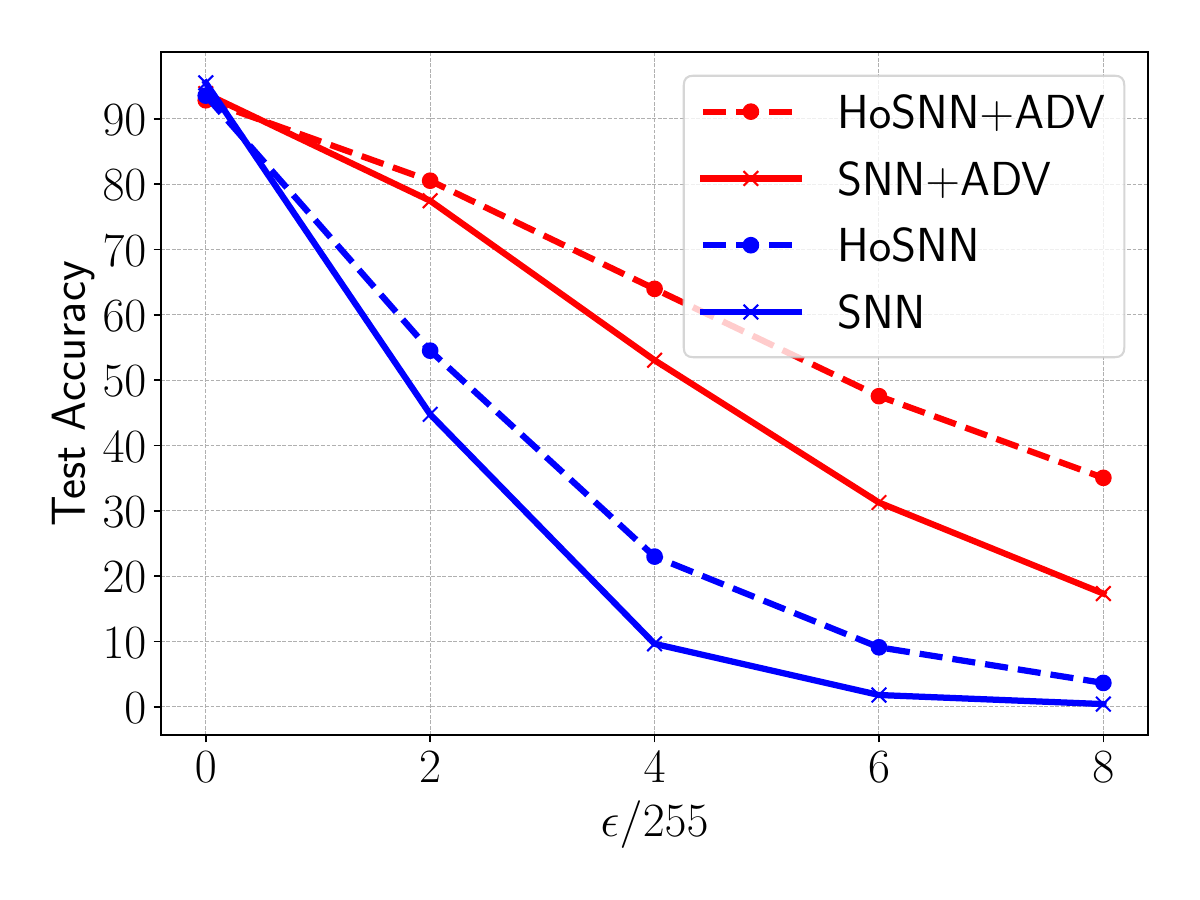}
                \caption{SVHN}
            \end{subfigure}
            \begin{subfigure}{0.245\linewidth}
                \includegraphics[width=\linewidth]{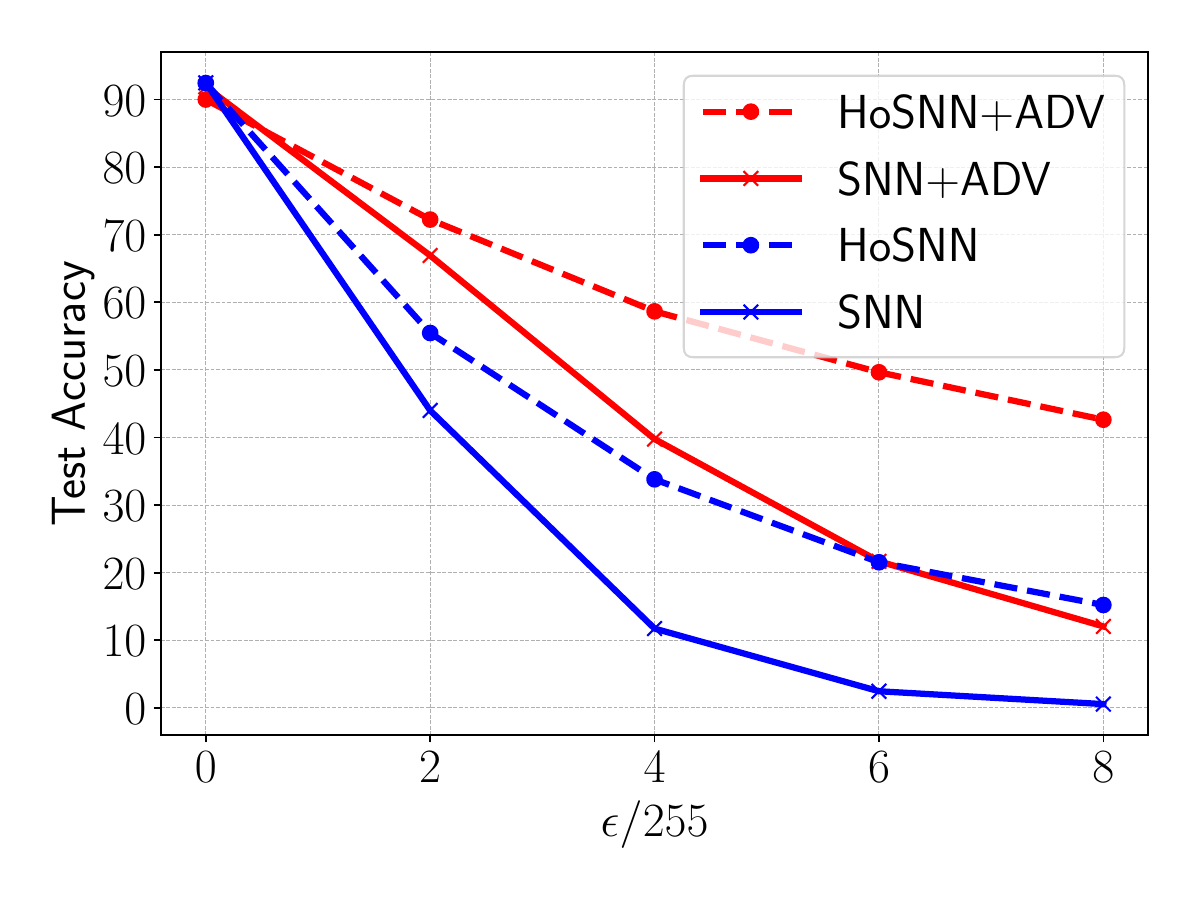}
                \caption{CIFAR-10}
            \end{subfigure}
            \begin{subfigure}{0.245\linewidth}
                \includegraphics[width=\linewidth]{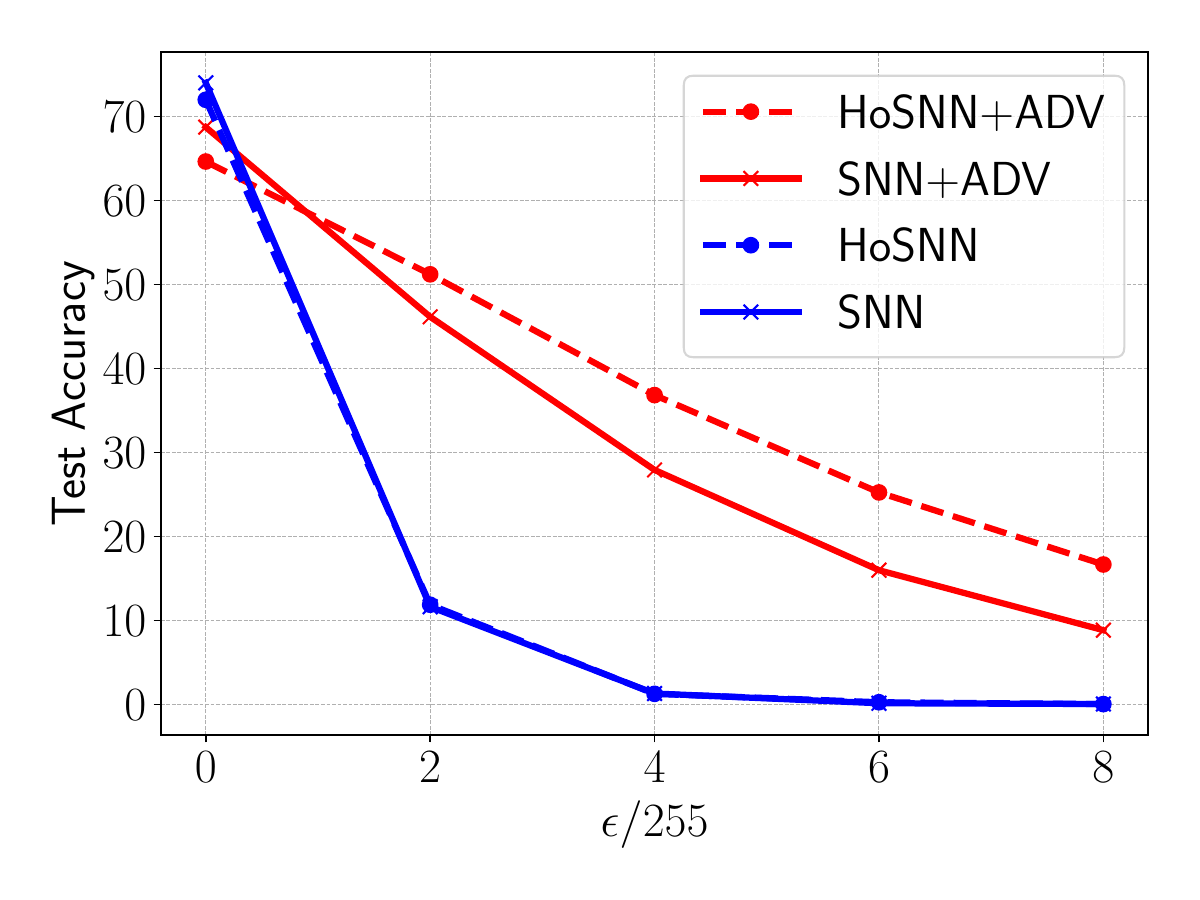}
                \caption{CIFAR-100}
            \end{subfigure}
            \caption{For Test(5). Performance of SNNs and HoSNNs as a function of white-box PGD7 attack intensity.}
            \label{fig:performance}
        \end{figure}

   \paragraph{For Test (5)} Since all gradient-based attacks work, there is no need to use random sampling methods. Therefore Test (5) passed obviously. Figure~\ref{fig:performance} shows model performance  with and without adversarial training (ADV) under white-box PGD7 attacks with varying intensities:  $\epsilon = 2, 4, 6, 8/255$. The proposed HoSNNs consistently outperform the SNNs in terms of model accuracy across all attack intensities and four datasets. 

\subsection{Comparison with other works}
\begin{wraptable}{r}{0.6\textwidth} 
    \centering
    \footnotesize
    \begin{tabular}{llllll}
        \toprule
        Data & Methods  & Clean & FGSM & PGD7 \\
        \hline    
        \multirow{3}{*}{\shortstack{CIFAR-\\10}} &
        \cite{ding2022snn} & 83.45 & 39.69 & 20.14  \\
        & \cite{ozdenizci2023adversarially} & $\mathbf{91.86}$ & 41.55 & 27.35  \\
        & Our work & 90.00 & $\mathbf{63.98}$ & $\mathbf{42.63}$   \\
        \hline
        \multirow{4}{*}{\shortstack{CIFAR-\\100}} 
        & \cite{ding2022snn} & 67.47 & 25.38 & 15.66  \\
        & \cite{ozdenizci2023adversarially} & $\mathbf{67.26}$ & 21.35 & 13.45  \\
         & Our work  & 64.64 & $\mathbf{26.97}$ & $\mathbf{16.66}$  \\
        \bottomrule
    \end{tabular}
    \caption{Comparison with others work. We use $\epsilon = 2$ FGSM adversarial training for CIFAR10 and $\epsilon = 4$ for CIFAR100. Adversarial examples are generated from the gradient of BPTT.}
    \label{tab:comparison}
\end{wraptable}
We compare our method against recent state-of-the-art defense methods under white-box FGSM and PGD7 attacks in Table~\ref{tab:comparison}. We tested all three methods on the same VGG7 network architecture, detailed in Appendix A.  We use $\epsilon = 2$ FGSM adversarial training for CIFAR10 and $\epsilon = 4$ for CIFAR100. Adversarial examples are generated from the gradient of BPTT. Our adversarially trained models achieved the highest defense accuracies.
\section{Discussions}
This paper presents the first work on biologically inspired homeostasis for enhancing adversarial robustness of spiking neural networks.   
Specifically, we propose a new TA-LIF model with a threshold adaptation mechanism and use TA-LIF neurons to construct inherently more robust HoSNN networks. Yet, there is room for future investigations including better trading off between model accuracy under clean and adversarial inputs.  
More broadly,  we recognize the vast and yet untapped potential of biological homeostasis in neural network research. The relationship between the properties of individual neurons and the overall performance of the network warrants further exploration.
\section{Acknowledgement}
This material is based upon work supported by the National Science Foundation under Grants No. 1948201 and No. 2310170.

\bibliography{main}
\bibliographystyle{tmlr}
\appendix
\section{Appendix}
\input{appendix}
\end{document}

%% file: math_commands.tex
\usepackage{amsmath,amsfonts,bm}

\def\eqref#1{equation~\ref{#1}}

\def\1{\bm{1}}

\DeclareMathAlphabet{\mathsfit}{\encodingdefault}{\sfdefault}{m}{sl}
\SetMathAlphabet{\mathsfit}{bold}{\encodingdefault}{\sfdefault}{bx}{n}



%% file: appendix.tex
\maketitle
We mainly present the derivation of the second-order dynamic equation of TA-LIF in~\ref{sup:dynamics}, dynamic stability analysis in~\ref{sup:stable} the detailed experimental setup in~\ref{sup:detail} and checklist for gradient obfuscation in~\ref{sup:gradient_obs}.

\section{Derivation of TA-LIF Dynamic Equation}
\label{sup:dynamics}
In this section, we derive the approximate second-order dynamic equations of the threshold-adapting leaky integrate-and-fire (TA-LIF) neurons and subsequently analyze them.

\subsection{LIF Dynamics}
To facilitate our discussion, let's commence by presenting the first-order dynamic equations of the LIF neuron \(i\) at time \(t\):
\begin{equation}
    \tau_m \frac{du_i(t)}{dt} = -u_i(t) + I_i(t) - \tau_m s_i(t)V_{th}
    \label{sup:eq1}
\end{equation}
The input current is defined as 
 
\begin{equation}
    I_i(t) = R~\sum_j w_{ij}a_j (t)
    \label{sup:eq2}
\end{equation}
The spiking behavior \(s_i(t)\) is defined as:
\begin{equation}
    s_i(t) = 
    \begin{cases} 
      +\infty & \text{if } u_i(t) \geq V_{th}\\
      0 & \text{otherwise}
    \end{cases} = \sum_f \delta(t-t_i^f)
    \label{sup:eq3}
\end{equation}
And the post-synaptic current dynamics are given by:

\begin{equation}
    \tau_s \frac{da_j(t)}{dt} = -a_j(t) + s_{j}(t)
    \label{sup:eq4}
\end{equation}
Where:
\begin{itemize}
    \item \( \tau_m \): Represents the membrane time constant.
    \item \( I_i(t) \): Denotes the input, which is the summation of the pre-synaptic currents.
    \item \( w_{ij} \): Stands for the synaptic weight from neuron \(j\) to neuron \(i\).
    \item \( a_j(t) \): Refers to the post-synaptic current induced by neuron \(j\) at time \(t\).
    \item \( V_{th}\): Is the static firing threshold.
    \item \( t_i^f \): Indicates the \(f\)-th spike time of neuron \(i\).
    \item \( \tau_s \): Is the synaptic time constant.
\end{itemize}

\subsection{Neural Dynamic Signature}
Let's begin by reviewing the definition of the Neural Dynamic Signature (NDS). Given a data instance \( x \) sampled from distribution \( \mathcal{D} \), the NDS of neuron \( i \), contingent upon the training set distribution \( \mathcal{D} \), can be represented as a temporal series vector $\bm{u^*_i}([\mathcal{D})$. Specifically, at time \( t \), it holds the value:
\begin{equation}
\label{sup:eq5}
u^*_i(t|\mathcal{D})  \triangleq \mathbb{E}_{x \sim \mathcal{D}}[u_i(t|x)],\ \text{for }  t \in [0,T]
\end{equation}
To derive the dynamics of NDS, we start by revisiting Equation \eqref{sup:eq1}, rewriting it with respect to  \( x\)
\begin{equation}
    \tau_m \frac{du_i(t| x)}{dt} = -u_i(t| x)+I_i(t| x)- \tau_m s_i(t| x)V_{th}
    \label{sup:eq6}
\end{equation}
For the convenience of dynamic analysis, we choose to approximate the discontinuous Dirac function term  \( s_i(t| x) \) with the average firing rate $r_i(\ x)$ of neuron $i$. The average firing rate is calculated by
\begin{equation}
    r_i(x)  \triangleq \int_0^T \frac{s_i(t| x)}{T} dt 
    \label{sup:eq7}
\end{equation}
We take this approximation to address discontinuity of Dirac funtion, as $r_i(x)$ and $s_i(t| x)$ have the same integral value over time: the number of neuron firings. Substituting Equation \eqref{sup:eq7} into Equation \eqref{sup:eq6} and computing the expectation on both sides, we have:
\begin{equation}
    \tau_m\mathbb{E}_{x \sim \mathcal{D}}[ \frac{du_i(t|x)}{dt}] = -\mathbb{E}_{x \sim \mathcal{D}}[u_i(t|x)] +\mathbb{E}_{x \sim \mathcal{D}}[I_i(t|x)] - \tau_m\mathbb{E}_{x \sim \mathcal{D}}[r_i(x)]V_{th}
    \label{sup:eq8}
\end{equation}
Here, we denote the average input current of neuron \( i \) over the entire dataset as:
\begin{equation}
I^*_i(t| \mathcal{D}) \triangleq \mathbb{E}_{x \sim \mathcal{D}}[I_i(t| x)] 
    \label{sup:eq9}
\end{equation}
 and the average spike frequency of neuron \( i \) over the entire dataset as: 
 \begin{equation}
r^*_i(\mathcal{D})\triangleq \mathbb{E}_{x \sim \mathcal{D}}[r_i(x)]
\label{sup:eq10}
\end{equation}
 With the definition from Equation \eqref{sup:eq5}, the dynamics of NDS can be expressed as:
\begin{equation}
    \tau_m \frac{d{u^*_i}(t| \mathcal{D})}{dt} = -u^*_i(t| \mathcal{D}) +I^*_i(t| \mathcal{D})- \tau_m r^*_i( \mathcal{D}) V_{th}
    \label{sup:eq11}
\end{equation}
As mentioned in the main text, we usually expect NDS to have precise semantic information of the distribution $\mathcal{D}$. So NDS should be obtained through a well-trained model with optimal weight parameter \( \bm{W^*} \). For clarity in the following sections, we use \( \bm{W^*} \) to represent the actually used NDS:
\begin{equation}
    \tau_m \frac{d{u^*_i}(t|  \bm{W^*}, \mathcal{D})}{dt} = -{u^*_i}(t| \bm{W^*}, \mathcal{D}) +I^*_i(t| \bm{W^*}, \mathcal{D})- \tau_m r^*_i( \bm{W^*}, \mathcal{D}) V_{th}
    \label{sup:eq12}
\end{equation}

\subsection{TA-LIF Dynamics}
In this section, we delve deeper into the dynamical equations governing the TA-LIF neuron and derive its second-order dynamic equation
\begin{equation}
    \tau_m \frac{du_i(t)}{dt} = -u_i(t) + I_i(t) - \tau_m s_i(t)V^i_{th}(t)
    \label{sup:eq13}
\end{equation}
The synaptic input \( I_i(t) \), the spike generation function \( s_i(t) \) and   post-synaptic current dynamics of TA-LIF are defined same as \eqref{sup:eq2}\eqref{sup:eq3}\eqref{sup:eq4}. For a specific network parameter \(  \bm{W} \) and a sample \( x' \) drawn from \( \mathcal{D'} \), the dynamic equation governing the threshold \( V^i_{th}(t) \) is:
\begin{equation}
    \frac{dV_{th}^{i}(t|\bm{W}, x')}{d t} = \theta_i e_i(t|\bm{W}, x'),
    \label{sup:eq14}
\end{equation}
where the error signal, utilizing the NDS as given in \eqref{sup:eq5}, is defined as:
\begin{equation}
    e_i(t|\bm{W}, x') \triangleq u_i(t|\bm{W}, x') - {u^*_i}(t| \bm{W^*}, \mathcal{D})
    \label{sup:eq15}
\end{equation}
Applying the continuity approximation for the Dirac function as per \eqref{sup:eq7} and incorporating the conditional dependency of \( \bm{W} \) and \( x' \), rewriting the dynamics for TA-LIF \eqref{sup:eq13} as:
\begin{equation}
    \tau_m \frac{du_i(t|\bm{W}, x')}{dt} = -u_i(t|\bm{W}, x')+I_i(t|\bm{W}, x') - \tau_m r_i(\bm{W}, x')V^i_{th}(t|\bm{W}, x')
    \label{sup:eq16}
\end{equation}
Subtracting \eqref{sup:eq12} from \eqref{sup:eq16} and employing \eqref{sup:eq15}, denoting
\begin{equation}
    \label{sup:eq17}
   \Delta I_i(t|\bm{W}, x') \triangleq I_i(t|\bm{W}, x')-I_i^*(t|\bm{W^*},\mathcal{D})  
\end{equation}
we derive the 1st-order dynamic of $e_i(t|\bm{W}, x')$
\begin{equation}
    \tau_m \frac{de_i(t|\bm{W}, x')}{dt} = -e_i(t|\bm{W}, x')
    +\Delta I_i(t|\bm{W}, x') - \tau_m[r_i(\bm{W}, x')V^i_{th}(t|\bm{W}, x') -r^*_i(\bm{W^*}, \mathcal{D})V_{th}]
    \label{sup:eq18}
\end{equation}
Differentiating \eqref{sup:eq18} with respect to time and utilizing the threshold dynamics from \eqref{sup:eq15}, we obtain:

\begin{equation}
    \tau_m \frac{d^2e_i(t|\bm{W}, x')}{dt^2} = -\frac{de_i(t|\bm{W}, x')}{dt} +\frac{d\Delta I_i(t|\bm{W}, x')}{dt} - {\tau_m}{\theta_i} r_i(\bm{W}, x')e_i(t|\bm{W}, x')
    \label{sup:eq19}
\end{equation}
For succinctness, we will omit dependencies  on \( \bm{W} \) and \( x' \), resulting in TA-LIF dynamics in the main text \eqref{sup:eq11}:

\begin{equation}
    \tau_m \frac{d^2e_i(t)}{dt^2} + \frac{de_i(t)}{dt} + r_i {\tau_m}{\theta_i} e_i(t) = \frac{d\Delta I_i(t)}{dt} 
    \label{sup:eq20}
\end{equation}
For the standard LIF neurons where \( \theta_i \rightarrow 0 \), the equation simplifies to:

\begin{equation}
    \tau_m \frac{d^2e_i(t)}{dt^2} + \frac{de_i(t)}{dt}  = \frac{d\Delta I_i(t)}{dt} 
    \label{sup:eq21}
\end{equation}

\section{Dynamic Stability Analysis}
\label{sup:stable}
In this section, we analyze the stability of \eqref{sup:eq20} and  \eqref{sup:eq21} to explore the influence of our dynamic threshold mechanism on the noise suppression ability of the TA-LIF neuron.

\subsection{BIBO Stability of Equation \eqref{sup:eq20}}

\textit{Characteristic Equation:} We first show the BIBO (Bounded Input, Bounded Output) stability \cite{ogata2010modern} of TA-LIF neurons based on \eqref{sup:eq20}. The characteristic equation of \eqref{sup:eq20} of non-silent ($r_i>0$) and non-degenerating ($\tau_m, \theta_i > 0 $) TA-LIF neurons is:
\begin{equation}
    \tau_m s^2 + s + r_i{\tau_m}{\theta_i} = 0
    \label{sup:eq22}
\end{equation}
and its roots are
\begin{equation}
s_{1,2} = \frac{-1 \pm \sqrt{ \Delta}}{2\tau_m},\ \Delta = 1 - 4 r_i {\tau_m^2}{\theta_i} 
\label{sup:eq23}
\end{equation}

\begin{itemize}
    \item For \( \Delta > 0 \): Both roots $s_{1,2}$ are real and negative.
    \item For \( \Delta = 0 \): There's a single negative real root.
    \item For \( \Delta < 0 \): Both roots are complex with negative real parts.
\end{itemize}
For a second-order system to be BIBO, the roots of its characteristic equation must be negative real or have negative parts, which is clearly the case for the TA-LIF model under the above three situations,  affirming the BIBO stability of \eqref{sup:eq20}.  The BIBO stability signifies that with the bounded driving input to system \eqref{sup:eq20}, the deviation of the TA-LIF neuron's membrane potential from its targeted NDS is also bounded, demonstrating the well control of the growth of error \( e_i(t) \).

\subsection{Stability of Equation \eqref{sup:eq20} Under White Noise}
To elucidate the dynamic characteristics of TA-LIF further, we adopt the prevalent method \cite{abbott1993asynchronous, brunel2000dynamics, gerstner2014neuronal, renart2004mean}, approximating \( \Delta I(t) \) with a Wiener process. This approximation effectively represents small, independent, and random perturbations. Hence, the driving force in equation \eqref{sup:eq20} $\frac{d\Delta I_i(t)}{dt}$ can be modeled by a Gaussian white noise \( F(t) \), leading to the well-established Langevin equation in stochastic differential equations theory \cite{kloeden1992stochastic, van1992stochastic, risken1996}:

\begin{equation}
 \label{sup:eq24}
     \frac{d^2e_i(t)}{dt^2} + \frac{1}{\tau_m} \frac{de_i(t)}{dt} + {r_i}{\theta_i} e_i(t) = F(t)
\end{equation}
Denoting $\langle\cdot\rangle$ as averaging over time, \( F(t) \) is a Gaussian white noise with variance $\sigma^2$ that satisfies:

\begin{equation}
    \label{sup:eq25}
    \left\{
    \begin{array}{ll}
        \langle F(t)\rangle & = 0, \\
        \left\langle F\left(t_1\right) F\left(t_2\right)\right\rangle & = \sigma^2 \delta\left(t_1-t_2\right), \\
        \left\langle F\left(t_1\right) F\left(t_2\right) \cdots F\left(t_{2 n+1}\right)\right\rangle & = 0, \\
        \left\langle F\left(t_1\right) F\left(t_2\right) \cdots F\left(t_{2 n}\right)\right\rangle & = \sum_{\text {all pairs }}\left\langle F\left(t_i\right) F\left(t_j\right)\right\rangle \cdot \left\langle F\left(t_k\right) F\left(t_l\right)\right\rangle \cdots
    \end{array}
    \right.
\end{equation}
where the sum has to be taken over all the different ways in which one can divide the $2 n$ time points $t_1 \cdots t_{2n}$ into $n$ pairs. Under this assumption \eqref{sup:eq25}, the solution of the Langevin equation\eqref{sup:eq24} is \cite{uhlenbeck1930theory, wang1945theory}:

\begin{equation}
\label{sup:eq26}
\left\langle[\Delta e_i(t)]^2\right\rangle=\frac{\tau_m \sigma^2}{2r_i \theta_i}\left[1-e^{\frac{- t}{2\tau_m}}\left(\cos \left(\omega_1 t\right)+\frac{\sin \left(\omega_1 t\right)}{2\omega_1\tau_m}\right)\right] = O(\sigma^2)
\end{equation}     
where $\Delta e_i(t) =  e_i(t)-  \langle e_i(t) \rangle$ and $\omega_1=\sqrt{{r_i}{\theta_i}-\frac{1}{4\tau^2_m}}$. While under the same assumptions \eqref{sup:eq25}, equation \eqref{sup:eq21} yields:

\begin{equation}
\label{sup:eq27}
\left\langle[\Delta e_i(t)]^2\right\rangle =\frac{\tau_m^2 \sigma^2}{2} \left(t-\tau_m + \tau_m e^{- t / \tau_m}\right) = O(\sigma^2t)
\end{equation}
Obviously, Gaussian white noise with zero mean \eqref{sup:eq25} leads $\langle e_i(t) \rangle = 0$, $\langle[\Delta e_i(t)]^2\rangle =  \langle e_i^2(t) \rangle $. Hence,

\begin{equation}
\label{sup:eq28}
    \frac{d\Delta I(t)}{dt} \sim F(t) \implies
    \begin{cases}
         &\langle e_i^2(t) \rangle_{LIF} = O(\sigma^2 t)\\ &\langle e_i^2(t) \rangle_{TA-LIF} = O(\sigma^2 )
    \end{cases}
\end{equation}
Significantly, the mean square error \(\langle e_i^2(t) \rangle_{TA-LIF}\) of the TA-LIF neuron remains bounded to \(O(\sigma^2)\) and doesn't increase over time. In contrast, under identical input perturbations, the mean square error \(\langle e_i^2(t) \rangle_{LIF}\) of the LIF neuron may grow unbounded with time, highlighting its potential susceptibility to adversarial attacks.

{
\subsection{A more intuitive explanation}
We would like to provide a more intuitive explanation of Eq~\ref{eq:err_dyn2} and Eq~\ref{eq:vth}. Notice that when $ \theta_i = 0$  according to  Eq~\ref{eq:vth}, TA-LIF no longer has a dynamically changing threshold and thus degenerates into standard LIF. Its second-order dynamics degenerates into:
$$
\tau_m \frac{d^2 e_i(t)}{d t^2}+\frac{d e_i(t)}{d t} = \varepsilon(t)
$$ 
An example to intuitively understand these two formulas is that they correspond to a driven damped oscillator in the physical world, that is, the movement of a ball connected to a spring under a driving force and damping environment. Considering $e_i(t)$ as the distance the ball deviates from the equilibrium position, we can understand the physical meaning of each term in the formula:
$\tau_m \frac{d^2 e_i(t)}{d t^2}$ describes the acceleration of the ball; $\frac{d e_i(t)}{d t}$ describes the frictional resistance of the ball;  $r_i \tau_m \theta_i e_i(t)$ describes the restoring force of the spring, which is the tendency of the spring to pull the ball toward the equilibrium position ($e_i(t) = 0$); $\varepsilon(t)$ describes the driving force exerted by the environment on the ball.}

{
With this explanation, we can easily understand the difference between the second-order dynamics of TA-LIF and LIF. The key term is $r_i \tau_m \theta_i e_i(t)$. For TA-LIF, it is like a small ball that is constantly pulled toward its initial equilibrium position by a spring when it is subjected to external disturbances: the result is that the ball will not deviate too far ($e_i(t)$ remains small). However, there is no such term in the dynamics of LIF, which causes $e_i(t)$ to become larger under external disturbances.
}

\section{Experiment Setting Details}
\label{sup:detail}
Our evaluation encompasses three benchmark datasets: FashionMNIST, SVHN, CIFAR10, and CIFAR100. For experimental setups, we deploy:

\begin{itemize}
    \item LeNet (32C5-P-64C5-P-1024-10) for FashionMNIST.
    \item VGGs (32C3-32C3-P-64C3-P-128C3-128C3-128-10) for SVHN. 
    \item VGGs (128C3-P-256C3-P-512C3-1024C3-512C3-1024-512-10) for CIFAR10.
    \item VGGs (128C3-P-256C3-P-512C3-1024C3-512C3-1024-1024-100) for  CIFAR100. 
\end{itemize}
Here, the notation 32C3 represents a convolutional layer with 32 filters of size \(3 \times 3\), and P stands for a pooling layer using \(2 \times 2\) filters. For the CIFAR10 and CIFAR100 datasets, we incorporated batch normalization layers and dropout mechanisms to mitigate overfitting and elevate the performance of the deep networks. In our experiments with FashionMNIST, SVHN, and CIFAR10, the output spike train of LIF neurons was retained to compute the kernel loss, as described in \cite{zhang2020temporal}. For CIFAR100, we directly employed softmax for performance.

For all HoSNN experiments, a preliminary training phase was carried out using an LIF SNN, sharing the same architecture, on the clean datasets to deduce the NDS.  Hyperparameters for LIF and TA-LIF neurons included a simulation time $T = 5$ , a Membrane Voltage Constant \(\tau_m = 5\), and a Synapse Constant \(\tau_s = 3\). For the TA-LIF results in the main text, we assigned \(\theta_i\) initialization values of 5 for FashionMNIST, SVHN, CIFAR10 and 3 for CIFAR100. All neurons began with an initial threshold of 1. The step function was approximated using \(\sigma(x) = \frac{1}{1+e^{-5x}}\), where $x = u(t) - V_{th}(t)$ and the BPTT learning algorithm was employed. For TA-ALIF neurons, the learning rate for \( \theta_i \) was set at $1/10$ of the rate designated for weights, ensuring hyperparameter stability during training. We also constrained \( \theta_i \) to remain non-negative during optimization, ensuring a possible transition from TA-LIF to LIF. For the generation of all gradient-based adversarial attacks, we assume that the attacker can know all $V_{th}(t)$ and use it in the gradient, even if it is generated dynamically and is not stored as network parameters. During training, we use $V_{th}(t)=V_{th}(0)$. We utilized the Adam optimizer with hyperparameters betas set to (0.9, 0.999), and the \(lr = 5 \times 10^{-4}\) with cosine annealing learning rate scheduler (\(T=\) epochs). We set batch size to 64 and trained for 200 epochs. All images were transformed into currents to serve as network input.  Our code is adapted from \cite{zhang2020temporal}. The experiment used four NVIDIA A100 GPUs. For CIFAR10 and CIFAR100, it took up to about 48 hours for adversarial training.

Regarding adversarial attack, we use an array of attack strategies, including FGSM, RFGSM, PGD, and BIM. For both CIFAR10 and CIFAR100, we allocated an attack budget with \( \epsilon=8/255 \). For iterative schemes like PGD, we set \(\alpha=2.5*\epsilon/steps\) and \(steps=7, 20, 40\), aligning with the recommendations in \cite{ding2022snn}. For the adversarial training phase, FGSM training was used with \(\epsilon\) values of 2/255 for CIFAR10 as per \cite{ding2022snn} and 4/255 for CIFAR100, following \cite{kundu2021hire}.

\begin{table}[h]
    \centering
    \begin{tabular}{lc}
        \hline 
        Items to identify gradient obfuscation & HoSNN \\
        \hline 
        (1) Single-step attack performs better compared  to iterative attacks & \checkmark \\
        (2) Black-box attacks perform better compared to white-box attacks & \checkmark \\
        (3) Increasing perturbation bound can't increase attack strength & \checkmark \\
        (4) Unbounded attacks can't reach $100 \%$ success  & \checkmark \\
        (5) Adversarial example can be found through  random sampling & \checkmark \\
        \hline
    \end{tabular}
    \caption{Checklist for gradient obfuscation}
\end{table}
\section{More Experiment Data}
\label{sup:gradient_obs}
Obfuscated gradients are a type of gradient masking that leads to a false sense of security when defending against adversarial examples \cite{athalye2018obfuscated, carlini2019evaluating}. Here we perform sanity checks including three obfuscated types and a checklist as per \cite{athalye2018obfuscated}. First, we examine three types of obfuscated gradients. Specifically, we use the same surrogate to train HoSNN from scratch and deliver attacks; decent clean accuracy and a smooth training process indicate that it's not \textbf{Shattered Gradient} with nonexistent or incorrect value. As our defense doesn't introduce any random factors, \textbf{Stochastic Gradient} is not applicable. Our method also doesn't include any multiple iterations of neural network
evaluation, so \textbf{Vanishing/Exploding Gradient} are also not applicable.

\begin{table}[h]
    \centering
    \begin{tabular}{lcc}
        \hline 
        Items to identify gradient obfuscation & HoSNN & Experiment \\
        \hline 
        (1) Single-step attack performs better compared  to iterative attacks & \checkmark & Fig~\ref{fig:fgmpgd_compare} and Table~\ref{tab:transposed_fmnist_other_results1}
\\
        (2) Black-box attacks perform better compared to white-box attacks & \checkmark & Fig~\ref{fig:wb_compare} and Table~\ref{tab:transposed_fmnist_other_results2} \\
        (3) Increasing perturbation bound can't increase attack strength & \checkmark & Fig~\ref{fig:ho_compare} and Table~\ref{tab:transposed_fmnist_other_results3}\\
        (4) Unbounded attacks can't reach $100 \%$ success  & \checkmark & Fig~\ref{fig:ho_compare} and Table~\ref{tab:transposed_fmnist_other_results3} \\
        (5) Adversarial example can be found through  random sampling & \checkmark & Fig~\ref{fig:performance}\\
        \hline
    \end{tabular}
    \caption{Checklist for gradient obfuscation}
\end{table}

\begin{table*}[h]
\centering
\begin{tabular}{|c|c|c|c|c|c|c|c|c|c|}
\hline
\textbf{Dataset} & \textbf{Method} & $\epsilon$ = \textbf{0} & \textbf{2} & \textbf{4} & \textbf{6} & \textbf{8} & \textbf{16} & \textbf{32} & \textbf{64} \\ \hline
\multirow{2}{*}{\textbf{FMNIST}} & WB-FGSM & 92.31 & 90.26 & 88.16 & 86.25 & 84.70 & 77.70 & 60.70 & 24.27 \\ \cline{2-10}
                                 & WB-PGD & 92.31 & 89.34 & 85.44 & 80.61 & 74.91 & 47.93 & 5.87 & 0.00 \\ \hline
\multirow{2}{*}{\textbf{SVHN}}   & WB-FGSM  & 92.84 & 83.96 & 75.85 & 68.45 & 61.78 & 43.45 & 28.64 & 20.79 \\ \cline{2-10}
                                 & WB-PGD & 92.84 & 80.53 & 63.98 & 47.56 & 35.06 & 10.40 & 2.00 & 0.21 \\ \hline
\multirow{2}{*}{\textbf{CIFAR-10}} & WB-FGSM  & 90.00 & 77.84 & 71.14 & 67.05 & 63.98 & 56.66 & 48.92 & 26.93 \\ \cline{2-10}
                                   & WB-PGD & 90.00 & 72.24 & 58.66 & 49.65 & 42.63 & 24.38 & 6.82 & 0.65 \\ \hline
\multirow{2}{*}{\textbf{CIFAR-100}} & WB-FGSM  & 64.63 & 52.78 & 42.12 & 33.79 & 26.97 & 12.47 & 4.01 & 2.68 \\ \cline{2-10}
                                     & WB-PGD & 64.64 & 51.20 & 36.82 & 25.23 & 16.65 & 1.99 & 0.00 & 0.00 \\ \hline
\end{tabular}
\caption{HoSNN accuracy data for Test1. We compared the performance under white-box FGSM and PGD attack. Our data shows that single-step attacks are strictly weaker than multi-step attacks.}
\label{tab:transposed_fmnist_other_results1}
\end{table*}

\begin{table*}[h]
\centering
\begin{tabular}{|c|c|c|c|c|c|c|c|c|c|}
\hline
\textbf{Dataset} & \textbf{Method} & $\epsilon$ = \textbf{0} & \textbf{2} & \textbf{4} & \textbf{6} & \textbf{8} & \textbf{16} & \textbf{32} & \textbf{64} \\ \hline
\multirow{2}{*}{\textbf{FMNIST}} & BB-PGD & 92.31 & 91.05 & 89.81 & 88.85 & 87.67 & 84.61 & 75.12 & 39.93 \\ \cline{2-10}
                                 & WB-PGD & 92.31 & 89.34 & 85.44 & 80.61 & 74.91 & 47.93 & 5.87 & 0.00 \\ \hline
\multirow{2}{*}{\textbf{SVHN}}   & BB-PGD & 92.84 & 89.31 & 86.10 & 82.68 & 78.49 & 57.02 & 25.96 & 8.02 \\ \cline{2-10}
                                 & WB-PGD & 92.84 & 80.53 & 63.98 & 47.56 & 35.06 & 10.40 & 2.00 & 0.21 \\ \hline
\multirow{2}{*}{\textbf{CIFAR-10}} & BB-PGD & 90.00 & 86.55 & 83.46 & 80.11 & 76.61 & 56.77 & 13.43 & 0.92 \\ \cline{2-10}
                                   & WB-PGD & 90.00 & 72.24 & 58.66 & 49.65 & 42.63 & 24.38 & 6.82 & 0.65 \\ \hline
\multirow{2}{*}{\textbf{CIFAR-100}} & BB-PGD & 64.64 & 61.68 & 59.22 & 56.67 & 54.06 & 42.08 & 16.90 & 1.82 \\ \cline{2-10}
                                    & WB-PGD & 64.64 & 51.20 & 36.82 & 25.23 & 16.65 & 1.99 & 0.00 & 0.00 \\ \hline
\end{tabular}
\caption{HoSNN accuracy data for Test2. We compared the performance under white-box PGD and black-box PGD attack. Our data shows that black-box attacks are strictly weaker than white-box attacks.}
\label{tab:transposed_fmnist_other_results2}
\end{table*}

\begin{table*}[h]
\centering
\begin{tabular}{|c|c|c|c|c|c|c|c|c|c|}
\hline
\textbf{Dataset} & \textbf{Method} & $\epsilon$ = \textbf{0} & \textbf{2} & \textbf{4} & \textbf{6} & \textbf{8} & \textbf{16} & \textbf{32} & \textbf{64} \\ \hline
\multirow{2}{*}{\textbf{FMNIST}} & SNN-WB-PGD & 92.92 & 78.87 & 58.03 & 42.27 & 30.54 & 2.71 & 0.00 & 0.00 \\ \cline{2-10}
                                 & HoSNN-WB-PGD & 92.31 & 89.34 & 85.44 & 80.61 & 74.91 & 47.93 & 5.87 & 0.00 \\ \hline
\multirow{2}{*}{\textbf{SVHN}}   & SNN-WB-PGD & 95.51 & 44.78 & 9.66 & 1.83 & 0.44 & 0.10 & 0.02 & 0.01 \\ \cline{2-10}
                                 &HoSNN-WB-PGD & 92.84 & 80.53 & 63.98 & 47.56 & 35.06 & 10.40 & 2.00 & 0.21 \\ \hline
\multirow{2}{*}{\textbf{CIFAR-10}} & SNN-WB-PGD & 92.47 & 44.00 & 11.73 & 2.46 & 0.56 & 0.01 & 0.00 & 0.00 \\ \cline{2-10}
                                   & HoSNN-WB-PGD & 90.00 & 72.24 & 58.66 & 49.65 & 42.63 & 24.38 & 6.82 & 0.65 \\ \hline
\multirow{2}{*}{\textbf{CIFAR-100}} & SNN-WB-PGD & 74.00 & 11.62 & 1.29 & 0.13 & 0.04 & 0.00 & 0.00 & 0.00 \\ \cline{2-10}
                                     & HoSNN-WB-PGD & 64.64 & 51.20 & 36.82 & 25.23 & 16.65 & 1.99 & 0.00 & 0.00 \\ \hline
\end{tabular}
\caption{HoSNN accuracy data for Test3\&4. We showed the performance of SNN and HoSNN under white-box PGD attack. Our data shows that increasing perturbation bound can increase attack strength and the accuracy can drop to 0 as the attack becomes stronger.}
\label{tab:transposed_fmnist_other_results3}
\end{table*}

\newpage

%% file: main.bbl
\begin{thebibliography}{82}
\providecommand{\natexlab}[1]{#1}
\providecommand{\url}[1]{\texttt{#1}}
\expandafter\ifx\csname urlstyle\endcsname\relax
  \providecommand{\doi}[1]{doi: #1}\else
  \providecommand{\doi}{doi: \begingroup \urlstyle{rm}\Url}\fi

\bibitem[Abbott \& Van~Vreeswijk(1993)Abbott and Van~Vreeswijk]{abbott1993asynchronous}
Larry~F Abbott and Carl Van~Vreeswijk.
\newblock Asynchronous states in networks of pulse-coupled oscillators.
\newblock \emph{Physical Review E}, 48\penalty0 (2):\penalty0 1483, 1993.

\bibitem[Akhtar \& Mian(2018)Akhtar and Mian]{akhtar2018threat}
Naveed Akhtar and Ajmal Mian.
\newblock Threat of adversarial attacks on deep learning in computer vision: A survey.
\newblock \emph{Ieee Access}, 6:\penalty0 14410--14430, 2018.

\bibitem[AllenInstitute(2018)]{Allen}
AllenInstitute.
\newblock \emph{Allen Cell Types Database, cell feature search.}
\newblock AllenInstitute, 2018.

\bibitem[Athalye et~al.(2018)Athalye, Carlini, and Wagner]{athalye2018obfuscated}
Anish Athalye, Nicholas Carlini, and David Wagner.
\newblock Obfuscated gradients give a false sense of security: Circumventing defenses to adversarial examples.
\newblock In \emph{International conference on machine learning}, pp.\  274--283. PMLR, 2018.

\bibitem[Bellec et~al.(2018)Bellec, Salaj, Subramoney, Legenstein, and Maass]{bellec2018long}
Guillaume Bellec, Darjan Salaj, Anand Subramoney, Robert Legenstein, and Wolfgang Maass.
\newblock Long short-term memory and learning-to-learn in networks of spiking neurons.
\newblock \emph{Advances in neural information processing systems}, 31, 2018.

\bibitem[Bellec et~al.(2020)Bellec, Scherr, Subramoney, Hajek, Salaj, Legenstein, and Maass]{bellec2020solution}
Guillaume Bellec, Franz Scherr, Anand Subramoney, Elias Hajek, Darjan Salaj, Robert Legenstein, and Wolfgang Maass.
\newblock A solution to the learning dilemma for recurrent networks of spiking neurons.
\newblock \emph{Nature communications}, 11\penalty0 (1):\penalty0 3625, 2020.

\bibitem[Bernard(1865)]{bernard1865introduction}
Claude Bernard.
\newblock \emph{Introduction {\`a} l'{\'e}tude de la m{\'e}decine exp{\'e}rimentale}.
\newblock Number~2. Bailli{\`e}re, 1865.

\bibitem[Biggio \& Roli(2018)Biggio and Roli]{biggio2018wild}
Battista Biggio and Fabio Roli.
\newblock Wild patterns: Ten years after the rise of adversarial machine learning.
\newblock In \emph{Proceedings of the 2018 ACM SIGSAC Conference on Computer and Communications Security}, pp.\  2154--2156, 2018.

\bibitem[Brunel(2000)]{brunel2000dynamics}
Nicolas Brunel.
\newblock Dynamics of sparsely connected networks of excitatory and inhibitory spiking neurons.
\newblock \emph{Journal of computational neuroscience}, 8:\penalty0 183--208, 2000.

\bibitem[Bu et~al.(2023)Bu, Ding, Hao, and Yu]{bu2023rate}
Tong Bu, Jianhao Ding, Zecheng Hao, and Zhaofei Yu.
\newblock Rate gradient approximation attack threats deep spiking neural networks.
\newblock In \emph{Proceedings of the IEEE/CVF Conference on Computer Vision and Pattern Recognition}, pp.\  7896--7906, 2023.

\bibitem[Carlini \& Wagner(2017)Carlini and Wagner]{carlini2017adversarialexampleseasilydetected}
Nicholas Carlini and David Wagner.
\newblock Adversarial examples are not easily detected: Bypassing ten detection methods, 2017.
\newblock URL \url{https://arxiv.org/abs/1705.07263}.

\bibitem[Carlini et~al.(2019)Carlini, Athalye, Papernot, Brendel, Rauber, Tsipras, Goodfellow, Madry, and Kurakin]{carlini2019evaluating}
Nicholas Carlini, Anish Athalye, Nicolas Papernot, Wieland Brendel, Jonas Rauber, Dimitris Tsipras, Ian Goodfellow, Aleksander Madry, and Alexey Kurakin.
\newblock On evaluating adversarial robustness, 2019.

\bibitem[Chakraborty et~al.(2018)Chakraborty, Alam, Dey, Chattopadhyay, and Mukhopadhyay]{chakraborty2018adversarial}
Anirban Chakraborty, Manaar Alam, Vishal Dey, Anupam Chattopadhyay, and Debdeep Mukhopadhyay.
\newblock Adversarial attacks and defences: A survey.
\newblock \emph{arXiv preprint arXiv:1810.00069}, 2018.

\bibitem[Chowdhury et~al.(2021)Chowdhury, Lee, and Roy]{chowdhury2021towards}
Sayeed~Shafayet Chowdhury, Chankyu Lee, and Kaushik Roy.
\newblock Towards understanding the effect of leak in spiking neural networks.
\newblock \emph{Neurocomputing}, 464:\penalty0 83--94, 2021.

\bibitem[Cooper(2008)]{cooper2008claude}
Steven~J Cooper.
\newblock From claude bernard to walter cannon. emergence of the concept of homeostasis.
\newblock \emph{Appetite}, 51\penalty0 (3):\penalty0 419--427, 2008.

\bibitem[Croce \& Hein(2020{\natexlab{a}})Croce and Hein]{croce2020reliable}
Francesco Croce and Matthias Hein.
\newblock Reliable evaluation of adversarial robustness with an ensemble of diverse parameter-free attacks.
\newblock In \emph{International conference on machine learning}, pp.\  2206--2216. PMLR, 2020{\natexlab{a}}.

\bibitem[Croce \& Hein(2020{\natexlab{b}})Croce and Hein]{croce2020reliableevaluationadversarialrobustness}
Francesco Croce and Matthias Hein.
\newblock Reliable evaluation of adversarial robustness with an ensemble of diverse parameter-free attacks, 2020{\natexlab{b}}.
\newblock URL \url{https://arxiv.org/abs/2003.01690}.

\bibitem[Deng et~al.(2020)Deng, Wu, Hu, Liang, Ding, Li, Zhao, Li, and Xie]{deng2020rethinking}
Lei Deng, Yujie Wu, Xing Hu, Ling Liang, Yufei Ding, Guoqi Li, Guangshe Zhao, Peng Li, and Yuan Xie.
\newblock Rethinking the performance comparison between snns and anns.
\newblock \emph{Neural networks}, 121:\penalty0 294--307, 2020.

\bibitem[Ding et~al.(2022)Ding, Bu, Yu, Huang, and Liu]{ding2022snn}
Jianhao Ding, Tong Bu, Zhaofei Yu, Tiejun Huang, and Jian Liu.
\newblock Snn-rat: Robustness-enhanced spiking neural network through regularized adversarial training.
\newblock \emph{Advances in Neural Information Processing Systems}, 35:\penalty0 24780--24793, 2022.

\bibitem[Ding et~al.(2024{\natexlab{a}})Ding, Pan, Liu, Yu, and Huang]{ding2024robust}
Jianhao Ding, Zhiyu Pan, Yujia Liu, Zhaofei Yu, and Tiejun Huang.
\newblock Robust stable spiking neural networks.
\newblock \emph{arXiv preprint arXiv:2405.20694}, 2024{\natexlab{a}}.

\bibitem[Ding et~al.(2024{\natexlab{b}})Ding, Yu, Huang, and Liu]{Ding_Yu_Huang_Liu_2024}
Jianhao Ding, Zhaofei Yu, Tiejun Huang, and Jian~K. Liu.
\newblock Enhancing the robustness of spiking neural networks with stochastic gating mechanisms.
\newblock \emph{Proceedings of the AAAI Conference on Artificial Intelligence}, 38\penalty0 (1):\penalty0 492--502, Mar. 2024{\natexlab{b}}.
\newblock \doi{10.1609/aaai.v38i1.27804}.
\newblock URL \url{https://ojs.aaai.org/index.php/AAAI/article/view/27804}.

\bibitem[El-Allami et~al.(2021)El-Allami, Marchisio, Shafique, and Alouani]{el2021securing}
Rida El-Allami, Alberto Marchisio, Muhammad Shafique, and Ihsen Alouani.
\newblock Securing deep spiking neural networks against adversarial attacks through inherent structural parameters.
\newblock In \emph{2021 Design, Automation \& Test in Europe Conference \& Exhibition (DATE)}, pp.\  774--779. IEEE, 2021.

\bibitem[Fawzi et~al.(2016)Fawzi, Moosavi-Dezfooli, and Frossard]{fawzi2016robustness}
Alhussein Fawzi, Seyed-Mohsen Moosavi-Dezfooli, and Pascal Frossard.
\newblock Robustness of classifiers: from adversarial to random noise.
\newblock \emph{Advances in neural information processing systems}, 29, 2016.

\bibitem[Ford et~al.(2019)Ford, Gilmer, Carlini, and Cubuk]{ford2019adversarial}
Nic Ford, Justin Gilmer, Nicolas Carlini, and Dogus Cubuk.
\newblock Adversarial examples are a natural consequence of test error in noise.
\newblock \emph{arXiv preprint arXiv:1901.10513}, 2019.

\bibitem[Furber et~al.(2014)Furber, Galluppi, Temple, and Plana]{furber2014spinnaker}
Steve~B Furber, Francesco Galluppi, Steve Temple, and Luis~A Plana.
\newblock The spinnaker project.
\newblock \emph{Proceedings of the IEEE}, 102\penalty0 (5):\penalty0 652--665, 2014.

\bibitem[Gerstner \& Kistler(2002)Gerstner and Kistler]{gerstner2002spiking}
Wulfram Gerstner and Werner~M Kistler.
\newblock \emph{Spiking neuron models: Single neurons, populations, plasticity}.
\newblock Cambridge university press, 2002.

\bibitem[Gerstner et~al.(2014)Gerstner, Kistler, Naud, and Paninski]{gerstner2014neuronal}
Wulfram Gerstner, Werner~M Kistler, Richard Naud, and Liam Paninski.
\newblock \emph{Neuronal dynamics: From single neurons to networks and models of cognition}.
\newblock Cambridge University Press, 2014.

\bibitem[Goodfellow et~al.(2014)Goodfellow, Shlens, and Szegedy]{goodfellow2014explaining}
Ian~J Goodfellow, Jonathon Shlens, and Christian Szegedy.
\newblock Explaining and harnessing adversarial examples.
\newblock \emph{arXiv preprint arXiv:1412.6572}, 2014.

\bibitem[Hao et~al.(2024)Hao, Bu, Shi, Huang, Yu, and Huang]{hao2024threaten}
Zecheng Hao, Tong Bu, Xinyu Shi, Zihan Huang, Zhaofei Yu, and Tiejun Huang.
\newblock Threaten spiking neural networks through combining rate and temporal information.
\newblock In \emph{The Twelfth International Conference on Learning Representations}, 2024.
\newblock URL \url{https://openreview.net/forum?id=xv8iGxENyI}.

\bibitem[Ilyas et~al.(2019{\natexlab{a}})Ilyas, Santurkar, Tsipras, Engstrom, Tran, and Madry]{ilyas2019adversarial}
Andrew Ilyas, Shibani Santurkar, Dimitris Tsipras, Logan Engstrom, Brandon Tran, and Aleksander Madry.
\newblock Adversarial examples are not bugs, they are features.
\newblock \emph{Advances in neural information processing systems}, 32, 2019{\natexlab{a}}.

\bibitem[Ilyas et~al.(2019{\natexlab{b}})Ilyas, Santurkar, Tsipras, Engstrom, Tran, and Madry]{ilyas2019adversarialexamplesbugsfeatures}
Andrew Ilyas, Shibani Santurkar, Dimitris Tsipras, Logan Engstrom, Brandon Tran, and Aleksander Madry.
\newblock Adversarial examples are not bugs, they are features, 2019{\natexlab{b}}.
\newblock URL \url{https://arxiv.org/abs/1905.02175}.

\bibitem[Imam \& Cleland(2020)Imam and Cleland]{imam2020rapid}
Nabil Imam and Thomas~A Cleland.
\newblock Rapid online learning and robust recall in a neuromorphic olfactory circuit.
\newblock \emph{Nature Machine Intelligence}, 2\penalty0 (3):\penalty0 181--191, 2020.

\bibitem[J{\"a}nig(2022)]{janig2022integrative}
Wilfrid J{\"a}nig.
\newblock \emph{The integrative action of the autonomic nervous system: neurobiology of homeostasis}.
\newblock Cambridge University Press, 2022.

\bibitem[Kang et~al.(2019)Kang, Sun, Hendrycks, Brown, and Steinhardt]{kang2019testing}
Daniel Kang, Yi~Sun, Dan Hendrycks, Tom Brown, and Jacob Steinhardt.
\newblock Testing robustness against unforeseen adversaries.
\newblock \emph{arXiv preprint arXiv:1908.08016}, 2019.

\bibitem[Kloeden et~al.(1992)Kloeden, Platen, Kloeden, and Platen]{kloeden1992stochastic}
Peter~E Kloeden, Eckhard Platen, Peter~E Kloeden, and Eckhard Platen.
\newblock \emph{Stochastic differential equations}.
\newblock Springer, 1992.

\bibitem[Krizhevsky(2009)]{krizhevsky2009learning}
Alex Krizhevsky.
\newblock Learning multiple layers of features from tiny images.
\newblock Technical report, Citeseer, 2009.

\bibitem[Kundu et~al.(2021)Kundu, Pedram, and Beerel]{kundu2021hire}
Souvik Kundu, Massoud Pedram, and Peter~A Beerel.
\newblock Hire-snn: Harnessing the inherent robustness of energy-efficient deep spiking neural networks by training with crafted input noise.
\newblock In \emph{Proceedings of the IEEE/CVF International Conference on Computer Vision}, pp.\  5209--5218, 2021.

\bibitem[Kurakin et~al.(2018)Kurakin, Goodfellow, and Bengio]{kurakin2018adversarial}
Alexey Kurakin, Ian~J Goodfellow, and Samy Bengio.
\newblock Adversarial examples in the physical world.
\newblock In \emph{Artificial intelligence safety and security}, pp.\  99--112. Chapman and Hall/CRC, 2018.

\bibitem[Lee et~al.(2020)Lee, Sarwar, Panda, Srinivasan, and Roy]{lee2020enabling}
Chankyu Lee, Syed~Shakib Sarwar, Priyadarshini Panda, Gopalakrishnan Srinivasan, and Kaushik Roy.
\newblock Enabling spike-based backpropagation for training deep neural network architectures.
\newblock \emph{Frontiers in neuroscience}, pp.\  119, 2020.

\bibitem[Li et~al.(2021)Li, Liu, Liu, Xu, Zhang, and Xie]{li2021understanding}
Tianlin Li, Aishan Liu, Xianglong Liu, Yitao Xu, Chongzhi Zhang, and Xiaofei Xie.
\newblock Understanding adversarial robustness via critical attacking route.
\newblock \emph{Information Sciences}, 547:\penalty0 568--578, 2021.

\bibitem[Li et~al.(2022)Li, Cui, Zhou, and Li]{li2022comparative}
Yanjie Li, Xiaoxin Cui, Yihao Zhou, and Ying Li.
\newblock A comparative study on the performance and security evaluation of spiking neural networks.
\newblock \emph{IEEE Access}, 10:\penalty0 117572--117581, 2022.

\bibitem[Liang et~al.(2022)Liang, Xu, Hu, Deng, and Xie]{liang2022toward}
Ling Liang, Kaidi Xu, Xing Hu, Lei Deng, and Yuan Xie.
\newblock Toward robust spiking neural network against adversarial perturbation.
\newblock \emph{Advances in Neural Information Processing Systems}, 35:\penalty0 10244--10256, 2022.

\bibitem[Lillicrap et~al.(2020)Lillicrap, Santoro, Marris, Akerman, and Hinton]{lillicrap2020backpropagation}
Timothy~P Lillicrap, Adam Santoro, Luke Marris, Colin~J Akerman, and Geoffrey Hinton.
\newblock Backpropagation and the brain.
\newblock \emph{Nature Reviews Neuroscience}, 21\penalty0 (6):\penalty0 335--346, 2020.

\bibitem[Liu et~al.(2024)Liu, Bu, Ding, Hao, Huang, and Yu]{liu2024enhancing}
Yujia Liu, Tong Bu, Jianhao Ding, Zecheng Hao, Tiejun Huang, and Zhaofei Yu.
\newblock Enhancing adversarial robustness in snns with sparse gradients.
\newblock \emph{arXiv preprint arXiv:2405.20355}, 2024.

\bibitem[Madry et~al.(2017)Madry, Makelov, Schmidt, Tsipras, and Vladu]{madry2017towards}
Aleksander Madry, Aleksandar Makelov, Ludwig Schmidt, Dimitris Tsipras, and Adrian Vladu.
\newblock Towards deep learning models resistant to adversarial attacks.
\newblock \emph{arXiv preprint arXiv:1706.06083}, 2017.

\bibitem[Marder \& Goaillard(2006)Marder and Goaillard]{marder2006variability}
Eve Marder and Jean-Marc Goaillard.
\newblock Variability, compensation and homeostasis in neuron and network function.
\newblock \emph{Nature Reviews Neuroscience}, 7\penalty0 (7):\penalty0 563--574, 2006.

\bibitem[Metzen et~al.(2017{\natexlab{a}})Metzen, Genewein, Fischer, and Bischoff]{metzen2017detecting}
Jan~Hendrik Metzen, Tim Genewein, Volker Fischer, and Bastian Bischoff.
\newblock On detecting adversarial perturbations.
\newblock \emph{arXiv preprint arXiv:1702.04267}, 2017{\natexlab{a}}.

\bibitem[Metzen et~al.(2017{\natexlab{b}})Metzen, Genewein, Fischer, and Bischoff]{metzen2017detectingadversarialperturbations}
Jan~Hendrik Metzen, Tim Genewein, Volker Fischer, and Bastian Bischoff.
\newblock On detecting adversarial perturbations, 2017{\natexlab{b}}.
\newblock URL \url{https://arxiv.org/abs/1702.04267}.

\bibitem[Modell et~al.(2015)Modell, Cliff, Michael, McFarland, Wenderoth, and Wright]{modell2015physiologist}
Harold Modell, William Cliff, Joel Michael, Jenny McFarland, Mary~Pat Wenderoth, and Ann Wright.
\newblock A physiologist's view of homeostasis.
\newblock \emph{Advances in physiology education}, 2015.

\bibitem[Moosavi-Dezfooli et~al.(2016)Moosavi-Dezfooli, Fawzi, and Frossard]{moosavi2016deepfool}
Seyed-Mohsen Moosavi-Dezfooli, Alhussein Fawzi, and Pascal Frossard.
\newblock Deepfool: a simple and accurate method to fool deep neural networks.
\newblock In \emph{Proceedings of the IEEE conference on computer vision and pattern recognition}, pp.\  2574--2582, 2016.

\bibitem[Mustafa et~al.(2019)Mustafa, Khan, Hayat, Goecke, Shen, and Shao]{mustafa2019adversarial}
Aamir Mustafa, Salman Khan, Munawar Hayat, Roland Goecke, Jianbing Shen, and Ling Shao.
\newblock Adversarial defense by restricting the hidden space of deep neural networks.
\newblock In \emph{Proceedings of the IEEE/CVF International Conference on Computer Vision}, pp.\  3385--3394, 2019.

\bibitem[Nadhamuni(2021)]{nadhamuni2021adversarial}
Kaveri Nadhamuni.
\newblock \emph{Adversarial Examples and Distribution Shift: A Representations Perspective}.
\newblock PhD thesis, Massachusetts Institute of Technology, 2021.

\bibitem[Neftci et~al.(2019)Neftci, Mostafa, and Zenke]{neftci2019surrogate}
Emre~O Neftci, Hesham Mostafa, and Friedemann Zenke.
\newblock Surrogate gradient learning in spiking neural networks: Bringing the power of gradient-based optimization to spiking neural networks.
\newblock \emph{IEEE Signal Processing Magazine}, 36\penalty0 (6):\penalty0 51--63, 2019.

\bibitem[Netzer et~al.(2011)Netzer, Wang, Coates, Bissacco, Wu, and Ng]{netzer2011reading}
Yuval Netzer, Tao Wang, Adam Coates, Alessandro Bissacco, Bo~Wu, and Andrew~Y Ng.
\newblock Reading digits in natural images with unsupervised feature learning.
\newblock 2011.

\bibitem[Ogata(2010)]{ogata2010modern}
Katsuhiko Ogata.
\newblock \emph{Modern control engineering fifth edition}.
\newblock 2010.

\bibitem[{\"O}zdenizci \& Legenstein(2023){\"O}zdenizci and Legenstein]{ozdenizci2023adversarially}
Ozan {\"O}zdenizci and Robert Legenstein.
\newblock Adversarially robust spiking neural networks through conversion.
\newblock \emph{arXiv preprint arXiv:2311.09266}, 2023.

\bibitem[Pei et~al.(2019)Pei, Deng, Song, Zhao, Zhang, Wu, Wang, Zou, Wu, He, et~al.]{pei2019towards}
Jing Pei, Lei Deng, Sen Song, Mingguo Zhao, Youhui Zhang, Shuang Wu, Guanrui Wang, Zhe Zou, Zhenzhi Wu, Wei He, et~al.
\newblock Towards artificial general intelligence with hybrid tianjic chip architecture.
\newblock \emph{Nature}, 572\penalty0 (7767):\penalty0 106--111, 2019.

\bibitem[Pennazio(2009)]{pennazio2009homeostasis}
Sergio Pennazio.
\newblock Homeostasis: a history of biology.
\newblock In \emph{Biology Forum/Rivista di Biologia}, volume 102, 2009.

\bibitem[Rabanser et~al.(2019)Rabanser, G{\"u}nnemann, and Lipton]{rabanser2019failing}
Stephan Rabanser, Stephan G{\"u}nnemann, and Zachary Lipton.
\newblock Failing loudly: An empirical study of methods for detecting dataset shift.
\newblock \emph{Advances in Neural Information Processing Systems}, 32, 2019.

\bibitem[Renart et~al.(2004)Renart, Brunel, and Wang]{renart2004mean}
Alfonso Renart, Nicolas Brunel, and Xiao-Jing Wang.
\newblock Mean-field theory of irregularly spiking neuronal populations and working memory in recurrent cortical networks.
\newblock \emph{Computational neuroscience: A comprehensive approach}, pp.\  431--490, 2004.

\bibitem[Risken(1996)]{risken1996}
H.~Risken.
\newblock \emph{The Fokker-Planck Equation: Methods of Solution and Applications}.
\newblock Springer, New York, 2 edition, 1996.
\newblock \doi{10.1007/978-3-642-61544-3_4}.

\bibitem[Selvaraju et~al.(2019)Selvaraju, Cogswell, Das, Vedantam, Parikh, and Batra]{Selvaraju_2019}
Ramprasaath~R. Selvaraju, Michael Cogswell, Abhishek Das, Ramakrishna Vedantam, Devi Parikh, and Dhruv Batra.
\newblock Grad-cam: Visual explanations from deep networks via gradient-based localization.
\newblock \emph{International Journal of Computer Vision}, 128\penalty0 (2):\penalty0 336–359, October 2019.
\newblock ISSN 1573-1405.
\newblock \doi{10.1007/s11263-019-01228-7}.
\newblock URL \url{http://dx.doi.org/10.1007/s11263-019-01228-7}.

\bibitem[Sharmin et~al.(2019)Sharmin, Panda, Sarwar, Lee, Ponghiran, and Roy]{sharmin2019comprehensive}
Saima Sharmin, Priyadarshini Panda, Syed~Shakib Sarwar, Chankyu Lee, Wachirawit Ponghiran, and Kaushik Roy.
\newblock A comprehensive analysis on adversarial robustness of spiking neural networks.
\newblock In \emph{2019 International Joint Conference on Neural Networks (IJCNN)}, pp.\  1--8. IEEE, 2019.

\bibitem[Sharmin et~al.(2020)Sharmin, Rathi, Panda, and Roy]{sharmin2020inherent}
Saima Sharmin, Nitin Rathi, Priyadarshini Panda, and Kaushik Roy.
\newblock Inherent adversarial robustness of deep spiking neural networks: Effects of discrete input encoding and non-linear activations.
\newblock In \emph{Computer Vision--ECCV 2020: 16th European Conference, Glasgow, UK, August 23--28, 2020, Proceedings, Part XXIX 16}, pp.\  399--414. Springer, 2020.

\bibitem[Shu et~al.(2020)Shu, Wu, Goldblum, and Goldstein]{shu2020prepare}
Manli Shu, Zuxuan Wu, Micah Goldblum, and Tom Goldstein.
\newblock Prepare for the worst: Generalizing across domain shifts with adversarial batch normalization.
\newblock 2020.

\bibitem[Silva \& Najafirad(2020)Silva and Najafirad]{silva2020opportunitieschallengesdeeplearning}
Samuel~Henrique Silva and Peyman Najafirad.
\newblock Opportunities and challenges in deep learning adversarial robustness: A survey, 2020.
\newblock URL \url{https://arxiv.org/abs/2007.00753}.

\bibitem[Simonyan \& Zisserman(2015)Simonyan and Zisserman]{simonyan2014very}
Karen Simonyan and Andrew Zisserman.
\newblock Very deep convolutional networks for large-scale image recognition.
\newblock In \emph{Proceedings of the international conference on learning representations (ICLR)}, 2015.

\bibitem[Szegedy et~al.(2013)Szegedy, Zaremba, Sutskever, Bruna, Erhan, Goodfellow, and Fergus]{szegedy2013intriguing}
Christian Szegedy, Wojciech Zaremba, Ilya Sutskever, Joan Bruna, Dumitru Erhan, Ian Goodfellow, and Rob Fergus.
\newblock Intriguing properties of neural networks.
\newblock \emph{arXiv preprint arXiv:1312.6199}, 2013.

\bibitem[Teeter et~al.(2018)Teeter, Iyer, Menon, Gouwens, Feng, Berg, Szafer, Cain, Zeng, Hawrylycz, et~al.]{teeter2018generalized}
Corinne Teeter, Ramakrishnan Iyer, Vilas Menon, Nathan Gouwens, David Feng, Jim Berg, Aaron Szafer, Nicholas Cain, Hongkui Zeng, Michael Hawrylycz, et~al.
\newblock Generalized leaky integrate-and-fire models classify multiple neuron types.
\newblock \emph{Nature communications}, 9\penalty0 (1):\penalty0 709, 2018.

\bibitem[Tram{\`e}r et~al.(2017)Tram{\`e}r, Kurakin, Papernot, Goodfellow, Boneh, and McDaniel]{tramer2017ensemble}
Florian Tram{\`e}r, Alexey Kurakin, Nicolas Papernot, Ian Goodfellow, Dan Boneh, and Patrick McDaniel.
\newblock Ensemble adversarial training: Attacks and defenses.
\newblock \emph{arXiv preprint arXiv:1705.07204}, 2017.

\bibitem[Tramèr(2022)]{tramèr2022detectingadversarialexamplesnearly}
Florian Tramèr.
\newblock Detecting adversarial examples is (nearly) as hard as classifying them, 2022.
\newblock URL \url{https://arxiv.org/abs/2107.11630}.

\bibitem[Turrigiano \& Nelson(2004)Turrigiano and Nelson]{turrigiano2004homeostatic}
Gina~G Turrigiano and Sacha~B Nelson.
\newblock Homeostatic plasticity in the developing nervous system.
\newblock \emph{Nature reviews neuroscience}, 5\penalty0 (2):\penalty0 97--107, 2004.

\bibitem[Uhlenbeck \& Ornstein(1930)Uhlenbeck and Ornstein]{uhlenbeck1930theory}
George~E Uhlenbeck and Leonard~S Ornstein.
\newblock On the theory of the brownian motion.
\newblock \emph{Physical review}, 36\penalty0 (5):\penalty0 823, 1930.

\bibitem[Van~Kampen(1992)]{van1992stochastic}
Nicolaas~Godfried Van~Kampen.
\newblock \emph{Stochastic processes in physics and chemistry}, volume~1.
\newblock Elsevier, 1992.

\bibitem[Wang \& Uhlenbeck(1945)Wang and Uhlenbeck]{wang1945theory}
Ming~Chen Wang and George~Eugene Uhlenbeck.
\newblock On the theory of the brownian motion ii.
\newblock \emph{Reviews of modern physics}, 17\penalty0 (2-3):\penalty0 323, 1945.

\bibitem[Woods \& Wilson(2013)Woods and Wilson]{woods2013information}
H~Arthur Woods and J~Keaton Wilson.
\newblock An information hypothesis for the evolution of homeostasis.
\newblock \emph{Trends in ecology \& evolution}, 28\penalty0 (5):\penalty0 283--289, 2013.

\bibitem[Wu et~al.(2018)Wu, Deng, Li, Zhu, and Shi]{wu2018spatio}
Yujie Wu, Lei Deng, Guoqi Li, Jun Zhu, and Luping Shi.
\newblock Spatio-temporal backpropagation for training high-performance spiking neural networks.
\newblock \emph{Frontiers in neuroscience}, 12:\penalty0 331, 2018.

\bibitem[Xiao et~al.(2017)Xiao, Rasul, and Vollgraf]{xiao2017fashion}
Han Xiao, Kashif Rasul, and Roland Vollgraf.
\newblock Fashion-mnist: a novel image dataset for benchmarking machine learning algorithms.
\newblock \emph{arXiv preprint arXiv:1708.07747}, 2017.

\bibitem[Xie et~al.(2017)Xie, Wang, Zhang, Ren, and Yuille]{xie2017mitigating}
Cihang Xie, Jianyu Wang, Zhishuai Zhang, Zhou Ren, and Alan Yuille.
\newblock Mitigating adversarial effects through randomization.
\newblock \emph{arXiv preprint arXiv:1711.01991}, 2017.

\bibitem[Xu et~al.(2022)Xu, Mahmood, Fang, Rathbun, Ding, and Wen]{xu2022securing}
Nuo Xu, Kaleel Mahmood, Haowen Fang, Ethan Rathbun, Caiwen Ding, and Wujie Wen.
\newblock Securing the spike: On the transferabilty and security of spiking neural networks to adversarial examples.
\newblock \emph{arXiv preprint arXiv:2209.03358}, 2022.

\bibitem[Zhang et~al.(2020)Zhang, Liu, Liu, Xu, Yu, Ma, and Li]{zhang2020interpreting}
Chongzhi Zhang, Aishan Liu, Xianglong Liu, Yitao Xu, Hang Yu, Yuqing Ma, and Tianlin Li.
\newblock Interpreting and improving adversarial robustness of deep neural networks with neuron sensitivity.
\newblock \emph{IEEE Transactions on Image Processing}, 30:\penalty0 1291--1304, 2020.

\bibitem[Zhang \& Li(2020)Zhang and Li]{zhang2020temporal}
Wenrui Zhang and Peng Li.
\newblock Temporal spike sequence learning via backpropagation for deep spiking neural networks.
\newblock \emph{Advances in Neural Information Processing Systems}, 33:\penalty0 12022--12033, 2020.

\end{thebibliography}
